\documentclass{article}

     \usepackage[preprint]{neurips_2022}

\usepackage[utf8]{inputenc} % allow utf-8 input
\usepackage[T1]{fontenc}    % use 8-bit T1 fonts
\usepackage{hyperref}       % hyperlinks
\usepackage{url}            % simple URL typesetting
\usepackage{booktabs}       % professional-quality tables
\usepackage{amsfonts}       % blackboard math symbols
\usepackage{nicefrac}       % compact symbols for 1/2, etc.
\usepackage{microtype}      % microtypography
\usepackage{xcolor}         % colors
\usepackage{amsmath}
\usepackage{amssymb}
\usepackage{amsthm}
\usepackage{bm}
\usepackage{bbm}
\usepackage{enumerate}
\usepackage{algorithm}
\usepackage[noend]{algpseudocode}
\usepackage{graphicx}
\usepackage{tikz}
\usepackage{xcolor}

\bibpunct{(}{)}{;}{a}{,}{,}

\theoremstyle{plain}
\newtheorem{theorem}{Theorem}
\newtheorem{lemma}{Lemma}
\newcommand{\ve}[1]{#1}
\newcommand{\define}[1]{\emph{#1}}
\newcommand{\nn}{\textsc{nn}}
\newcommand{\parameter}{\theta}

\newcommand{\diam}{\mathrm{diam}}

\title{A Multilabel Classification Framework for Approximate Nearest Neighbor Search}

\author{%
  Ville Hyv\"onen\\
  Department of Computer Science\\
  University of Helsinki\\
  Helsinki Institute for Information Technology (HIIT)\\
  \texttt{ville.o.hyvonen@helsinki.fi} \\
\And
  Elias J\"a\"asaari\\
  Machine Learning Department\\
  Carnegie Mellon University\\
  \texttt{ejaeaesa@cs.cmu.edu} \\
\And
  Teemu Roos\\
  Department of Computer Science\\
  University of Helsinki\\
  Helsinki Institute for Information Technology (HIIT)\\
  \texttt{teemu.roos@cs.helsinki.fi} \\
}

\begin{document}

\maketitle

\begin{abstract}

Both supervised and unsupervised machine learning algorithms have been used to learn partition-based index structures for approximate nearest neighbor (ANN) search. Existing supervised algorithms formulate the learning task as finding a partition in which the nearest neighbors of a training set point belong to the same partition element as the point itself, so that the nearest neighbor candidates can be retrieved by naive lookup or backtracking search. We formulate candidate set selection in ANN search directly as a multilabel classification problem where the labels correspond to the nearest neighbors of the query point, and interpret the partitions as partitioning classifiers for solving this task. Empirical results suggest that the natural classifier based on this interpretation leads to strictly improved performance when combined with any unsupervised or supervised partitioning strategy. We also prove a sufficient condition for consistency of a partitioning classifier for ANN search, and illustrate the result by verifying this condition for chronological $k$-d trees.

\end{abstract}

\section{Introduction}

Approximate nearest neighbor (ANN) search is a fundamental algorithmic problem. There is a large body of literature on ANN search spanning several research communities, including the machine learning community. Specifically, space-partitioning data structures---such as space-partitioning trees \citep{Friedman1976,Muja2014,Dasgupta2015} and data-dependent hash tables \citep{Indyk1998,Datar2004,Weiss2009}---are machine learning methods commonly used for ANN search. 

In this article, we propose an intuitive theoretical framework for partition-based ANN search. In particular, we formulate the candidate set selection directly as a multilabel classification problem where the labels represent the indices of the nearest neighbors of the query point. This formulation suggests that the performance of space-partitioning data structures can be improved by using them in a theoretically justified fashion as partitioning classifiers \citep[Chapter 21]{devroye2013probabilistic} instead of searching them under the earlier lookup-based paradigm. Our classification framework also enables applying general purpose classifiers---such as a multilabel random forest---directly as an index structure for ANN search.

We start by reviewing the relevant background on ANN search and multilabel classification (Sec.~\ref{sec:background}), and formulating candidate set selection in ANN search as a multilabel classification task (Sec.~\ref{sec:ann2multilabel}).
In Sec.~\ref{sec:partitioning_classifiers} we define the natural (partitioning) classifier for the general multilabel classification task.
%In Sec.~\ref{sec:related_work}, we review the earlier supervised partitioning methods (e.g., \citep{Cayton2008,Norouzi2011,Dong2020}) for ANN search.
In Sec. \ref{sec:candidate_set_selection} we show that interpreting the earlier lookup-based candidate set selection methods in the multilabel classification framework of Sec. \ref{sec:ann2multilabel} suggests that they define a suboptimal classifier. Our multilabel formulation also enables us to consider asymptotics in the standard statistical learning framework: we establish a sufficient condition for consistency of a partitioning classifier for ANN search (Sec.~\ref{sec:theory_sufficient_condition}). As a concrete example (Sec. \ref{sec:theory_kd_tree}), we verify this condition for the chronological $k$-d tree \citep{Bentley1975} that was the first data structure proposed for accelerating nearest neighbor search. To empirically validate the proposed framework, we show that using a natural classifier that is aligned with the ANN task  in conjunction with space-partitioning data structures proposed in the literature leads to strictly improved empirical performance compared to the earlier lookup-based candidate set selection methods (Sec.~\ref{sec:experiments}).

% We start by reviewing the relevant background on ANN search and multilabel classification (Sec.~\ref{sec:background}), and then formulate candidate set selection in ANN search as a multilabel classification task (Sec.~\ref{sec:ann2multilabel}). In Sec.~\ref{sec:related_work}, we explain how our proposed framework differs from earlier supervised partitioning methods (e.g., \citep{Cayton2008,Norouzi2011,Dong2020}). Our multilabel formulation also enables us to consider asymptotics in the standard statistical learning framework: we establish a sufficient condition for consistency of a partitioning classifier for ANN search (Sec.~\ref{sec:theory_sufficient_condition}). As a concrete example (Sec. \ref{sec:theory_kd_tree}), we verify this condition for the chronological $k$-d tree \citep{Bentley1975} that was the first data structure proposed for accelerating nearest neighbor search. To validate the theoretical findings, we show that using a natural classifier that is aligned with the ANN task  in conjunction with space-partitioning data structures proposed in the literature leads to strictly improved empirical performance compared to the earlier lookup-based candidate set selection methods (Sec.~\ref{sec:experiments}).

\section{Background and notation}
\label{sec:background}

% In this section, we review conventional formulations of ANN search and multilabel classification. 

% We will also define the necessary notation for the remainder of the paper.

\subsection{Approximate nearest neighbor search}
\label{sec:ANN_search}

Let the corpus points $\{c_j\}_{j=1}^m$ and the query point $x$ be vectors in $\mathbb{R}^d$. We call the $k$ corpus points that are closest\footnote{In what follows, we assume that the ties are broken uniformly at random so that the query point always has exactly $k$ nearest neighbors.} to the query point $x$ its \define{$k$ nearest neighbors} and denote the set of their indices by
\begin{equation}
\label{eq:nn_definition}
\nn_k(x) := \underset{j=1, \dots, m}{k\mathrm{-argmin}} \,\, \|x - c_j\|,
\end{equation}
where the notation ${k\mathrm{-argmin}} \,\, f$ means the set of $k$ values for which the function $f$ has the smallest values, and $\|\cdot\|$ is the Euclidean distance. Other metrics, or---more generally---dissimilarity measures can also be used to define the nearest neighbors. 

The trivial solution to the problem of finding the nearest neighbors $\nn_k(x)$ of the query point $x$ is to compute the distances to all the corpus points and sort these distances. However, when the dimensionality of the data is high and the corpus size is large, this brute force solution is often too slow if the application requires fast response times. The first data structure proposed for speeding up nearest neighbor search was the $k$-d tree \citep{Bentley1975}. However, for high-dimensional data, a $k$-d tree is not faster than the brute force approach for exact nearest neighbor search because of the well-known \textit{curse of dimensionality} that affects the non-parametric statistical methods---including partitioning methods for nearest neighbor search---in general \citep{lee1977worst}. Although the query speed of index structures for \emph{exact} nearest neighbor search degrades when the dimensionality of the data increases so that they are not an improvement on the brute force approach, this problem can be mitigated by allowing an approximate solution. This is why in modern high-dimensional applications \textit{approximate nearest neighbor} (ANN) search is typically used when a fast solution to the nearest neighbor problem is required.

Algorithms for ANN search can be divided into three categories: graphs~\citep{Malkov2014,Malkov2018,Iwasaki2018,Baranchuk2019}, quantization~\citep{Jegou2010,Johnson2019,Sablayrolles2019}, and space-partitioning methods. In this article, we consider space-partitioning methods that can be further divided into tree-based \citep{Muja2014,Dasgupta2015,Jaasaari2019} and hashing-based \citep{Datar2004,Aumuller2019b,gong2020idec} algorithms that use trees and hash tables, respectively, as index structures.

Space-partitioning algorithms for ANN search use an index structure to select a \define{candidate set} $S(x) \subset \{1, \dots, m\}$ of potential nearest neighbors. They then calculate the exact distances between the points in the candidate set and the query point, and return the $k$ nearest points as the approximate nearest neighbors. These algorithms will correctly retrieve a nearest neighbor $j\in\nn_k(x)$ if and only if it belongs to the candidate set. Thus, the \define{recall} of a space-partitioning algorithm can be written as $\mathrm{Rec}(S(\ve{x})) := \frac{1}{k}| \nn_k(\ve{x}) \cap S(x)|$, where we denote the number of elements of the set $A$ by $|A|$. The performance of an approximate nearest neighbor algorithm is typically measured by its average \emph{recall-query time tradeoff} (see e.g. \cite{Aumuller2019a} or \cite{Li2019})---i.e., the average query time required to reach a certain average recall level on a set of test queries.

\subsection{Multilabel classification}
\label{sec:multilabel_classification}

Consider a standard multi-label classification problem with $m$ labels. Let $X \in \mathbb{R}^d$ be a random variable and let $L(X) \subseteq \{1, \dots, m\}$ be the corresponding label set. Equivalently, the output variable can be presented in binary encoding by defining $Y \in \{0,1\}^m$ as an $m$-bit random vector, where
\begin{equation}
\label{eq:output_variable}
Y_{j} = \begin{cases}
1, \quad &\mathrm{if} \,\, j \in L(X), \\
0 &\mathrm{otherwise}.
\end{cases}
\end{equation}
A multilabel classifier is an $m$-component function $g = (g_1, \dots, g_m):\mathbb{R}^d \rightarrow \{0,1\}^m$ that attaches a label set to the value of the input variable $X$. Denote the training set that is assumed to be an i.i.d. sample from the distribution of the pair $(X,Y)$ by $D_n := \{(X_i, Y_i)\}_{i=1}^n$. When the classifier $g:\mathbb{R}^d \times \{\mathbb{R}^d \times \{0,1\}^m\}^n \rightarrow \{0,1\}^m$ is learned from the training set of size $n$, we denote it by $g^{(n)}(x) := g(x, D_n)$. When the training set $D_n$ is considered a random variable, the classifier $g^{(n)}$ also becomes a random function.% since it is a function of $D_n$. to emphasize that it is the function of both the input variable and the training set

The performance of the classifier is measured by a loss function $L:\{0,1\}^m \times \{0,1\}^m \rightarrow \mathbb{R}$, and the objective is to minimize the risk $\mathcal{R}(g) := E[L(g(X),Y)]$. This risk is lower-bounded by the \emph{Bayes risk} $\mathcal{R}^* = {\inf}_g \,\, \mathcal{R}(g)$, the minimizer of which is called the \emph{Bayes classifier}.

The Bayes classifier for many common multilabel loss functions---such as Hamming loss, ranking loss, precision, recall, and $F$-measures---is obtained by thresholding the conditional label probabilities $\eta_j(x) := P\{Y_j = 1 \,|\, X = x\}$ \citep{dembczynski2010bayes,koyejo2015consistent}. This justifies the standard plug-in approach of first estimating the conditional label probabilities\footnote{More generally, instead of the conditional label probability estimates $\hat{\eta}_1(x), \dots, \hat{\eta}_m(x)$, any score function values $s_1(x), \dots, s_m(x)$ for the labels can be learned and thresholded to make the classification decision. While we will present all the results only for the version of the plug-in classifier that uses the probability estimates, they readily generalize to the version of the plug-in classifier that uses the score function values.} by $\hat{\eta}_1(x), \dots, \hat{\eta}_m(x)$, and then defining the \emph{plug-in classifier} as 
\begin{equation}
\label{eq:plug_in_classifier}
g_j^{(n)}(x) := \begin{cases}
1, \quad &\mathrm{if} \,\, \hat{\eta}_j(x) > \tau \\
0, &\mathrm{otherwise},
\end{cases}
\end{equation}
where $\tau \in [0,1]$; equivalently, the plug-in classifier can be written as the estimate of the label set $L(x)$ as $\hat{L}(x) := \{j \in \{1, \dots, m\} \,:\, \hat{\eta}_j(x) > \tau\}$.

The multilabel classification problem is often solved by reducing it to a series of binary or multiclass classification problems, and estimating the conditional label probabilities $\eta_j(x)$ under this model. (see, e.g., \citet{menon2019multilabel} for a discussion of different reduction methods). In what follows, we will employ the \emph{pick-all-labels} (PAL) reduction \citep{reddi2019stochastic} where we separate each label $l \in L(x_i)$ of the training set point $x_i$ into a multi\emph{class} (but single-label) training instance $(x_i, l)$, and fit the classifier to this modified training set by minimizing a multiclass loss function.

\section{Candidate set selection as a multilabel classification problem}
\label{sec:ann2multilabel}

Equipped with the above definitions, we are now in a position to formalize  %non-probabilistic 
candidate set selection in ANN search described in Sec.~\ref{sec:ANN_search} 
%in the probabilistic framework 
as an instance of the multilabel classification problem described in Sec.~\ref{sec:multilabel_classification}. 

In the classical formulation of ANN search, the input--output pair is defined as $(x,\nn_k(x))$. It is straightforward to observe that this is essentially an instance of the multilabel classification problem where $\nn_k(x)$---i.e., the set of indices of the $k$ nearest neighbors of the query point---is the label set $L(x)$. Assuming that the values of $x$ are i.i.d. draws from the distribution of the random variable $X$ (the \emph{query distribution}), the objective is to predict the value of the random variable $Y$ (defined by \eqref{eq:output_variable} with $\nn_k(X)$ as a label set) given the value of the random variable $X$. A distinctive property of this classification problem, which follows from the definition of the ANN task, is that the size of the label set $|L(x)| = \sum_{j=1}^m y_j  = k$ is constant for all queries.

Since the labels $\{1, \dots, m\}$ correspond to the indices of the corpus points, the classification decision \eqref{eq:plug_in_classifier} where the probability estimates are thresholded corresponds to candidate set selection, and the estimated label set $\hat{L}(x)$ corresponds to the candidate set $S(x)$. 

If no additional training data is available, the corpus itself can be used as a training set. More precisely, in this case we interpret $\{c_j\}_{j=1}^m$ as a sample from the query distribution, compute the $k$ nearest neighbors of the corpus points, and then use $\{(c_j, y_j)\}_{j=1}^m$ as a training set. Note that in this case $y_{jj} = 1$ for each $j = 1, \dots, m$ since each corpus point is the nearest neighbor of itself.

\section{Partitioning classifiers}
\label{sec:partitioning_classifiers}

In this section, we first give a general definition of the partitioning classifier. We then define the \emph{natural classifier}---that is a special case of the partitioning classifier---for the single partition and for the ensemble of partitions for the general multilabel classification problem described in Sec. \ref{sec:multilabel_classification}. 

\emph{Partitioning classifier} is a general term for a classifier that is based on learning a partition of the instance space and whose classification decision is based on the labels of the training set points that belong to the same partition element as the query point. Partitioning classifiers can be divided into two categories depending on whether the partition is flat or recursive. There is a vast literature on recursive partitioning classifiers (i.e., classification trees), and gradient boosted trees \citep{Friedman2000,Friedman2001} are one of the most widely used and efficient classifiers \citep{Chen2016}. Flat partitions are more typically used for density estimation \citep{kontkanen2007mdl,lopez2013histogram,cui2021gbht}, but they have also been used for classification \citep{lugosi1996consistency,mcallester2003concentration}.

Denote by $\mathcal{P} = \{R_1, R_2, \dots, R_L\}$ the partition of $\mathbb{R}^d$, i.e., a collection of disjoint sets for which $\bigcup_{l=1}^L R_l = \mathbb{R}^d$. Denote the structure function that maps the query point to the index of the  partition element it belongs into by $q:\mathbb{R}^d \rightarrow \{1,2, \dots, L\}$. When the partition is learned from the training data, we denote it by $\mathcal{P}^{(n)} = \pi(D_n)$, where $\pi(D_n)$ is a partitioning rule that associates the training set with a partition of $\mathbb{R}^d$.

\paragraph{Natural classifier for a single partition.} Partitioning classifiers use the training set twice: first, to learn the partition $\mathcal{P}^{(n)} = \pi(D_n)$, and second, to classify the query point using the training set points that belong to the same partition element $R_{q(x)}$ as the query point $x$. In the multiclass classification, the conditional label probabilities can be estimated in a natural fashion by the observed label proportions
\begin{equation}
\label{eq:conditional_probability_estimate}
\hat{\eta}_j(x) = \frac{1}{N_{q(x)}} \sum_{i \,: \, x_i \in R_{q(x)}} y_{ij},
\end{equation}
where $N_{q(x)} := |\{i \,: \, x_i \in R_{q(x)}\} |$ denotes the number of training set points in the same partition element as the query point. This standard practice can be motivated by noting that these observed label proportions are the maximum likelihood estimates of the piecewise constant multinomial model where the conditional label probabilities are constants at each of the partition elements.

In the multilabel case, the estimation of the conditional label probabilities by \eqref{eq:conditional_probability_estimate}
can be motivated via the PAL reduction under which the observed label proportions in $R_{q(x)}$ are proportional to the maximum likelihood estimates of the piecewise constant multinomial model. To classify the query point, the conditional probability estimates \eqref{eq:conditional_probability_estimate} are plugged into \eqref{eq:plug_in_classifier}, i.e., the query point is assigned into all the classes whose probability estimate is greater than or equal to the value of the threshold parameter $\tau$. We call this partitioning classifier %that uses the observed label proportions as conditional label estimates 
the \emph{natural classifier}.

% \paragraph{Natural classifier for a single partition.} Partitioning classifiers use the training set twice: first, to learn the partition $\mathcal{P}^{(n)} = \pi(D_n)$, and second, to classify the query point using the training set points that belong to the same partition element $R_{q(x)}$ as the query point $x$. When a partitioning classifier is used for binary or multiclass classification, the query point $x$ is typically classified into the majority class of the training set points of the partition element $R_{q(x)}$ it belongs to. In multilabel classification, the concept of majority voting is not well-defined since the query point can belong to more than one class. Instead, we typically estimate the conditional probabilities of the labels by the observed label proportions
% \begin{equation}
% \label{eq:conditional_probability_estimate}
% \hat{\eta}_j(x) = \frac{1}{N_{q(x)}} \sum_{i \,: \, x_i \in R_{q(x)}} y_{ij},
% \end{equation}
% where $N_{q(x)} := |\{i \,: \, x_i \in R_{q(x)}\} |$ is the number of training set points in the same partition element as the query point. To classify the query point, the conditional probability estimates are plugged into \eqref{eq:plug_in_classifier}, i.e., the query point is assigned into all the classes whose probability estimate is greater than or equal to the value of the threshold parameter $\tau$. We call this partitioning classifier %that uses the observed label proportions as conditional label estimates 
% the \emph{natural classifier}. 

\paragraph{Natural classifier for an ensemble of partitions.} When a collection of partitions $\{\mathcal{P}^{(n)}_t\}_{t=1}^T$, where $\mathcal{P}_t^{(n)} := \{R_1^{(t)}, \dots, R_{L_t}^{(t)}\}$, is used as a classifier---such as in random forests \citep{Ho1998,Breiman2001}---the contributions of the partitions are aggregated. In this article, we consider the most straightforward aggregation method where the conditional probability estimates are obtained as averages of the conditional probability estimates of the individual partitions:
\begin{equation}
\label{eq:natural_classifier_ensemble_og}
\hat{\eta}_j(x) =  \frac{1}{T}\sum_{t=1}^T \hat{\eta}^{(t)}_j(x).
\end{equation}
The estimate of the single partition $\hat{\eta}_j^{(t)}(x)$ is defined as in \eqref{eq:conditional_probability_estimate} for the corresponding partition $\mathcal{P}^{(n)}_t$ and the corresponding structure function $q_t$. 

\section{Related work}
\label{sec:related_work}

The most directly relevant earlier literature consists of studies that learn space-partitioning index structures for ANN search using supervised information. The idea of optimising the index structure for the particular query distribution was first presented by \citet{maneewongvatana2001analysis}, and later extended by \citet{Cayton2008} who formulate ANN search as a supervised learning problem and propose a tree-based and a hashing-based algorithm for solving it. More recently, many supervised \emph{learning to hash}-methods, such as minimal loss hashing~\citep{Norouzi2011}, LDA hashing~\citep{Strecha2011}, and kernel-based supervised hashing~\citep{Liu2011}, have also been proposed for ANN search (see, e.g., \cite{Wang2015} or~\cite{Wang2017} for a survey). 

However, the earlier supervised methods pose the supervised learning problem in an indirect fashion. This is because they, like the earlier unsupervised methods, select the candidate set using a method which we call \emph{lookup search}\footnote{Often, lookup search is combined with a priority queue guided backtracking search in which the query point is routed into more than one element in a partition, and all the corpus points that belong to these partition elements are then selected into the candidate set. This technique is called \emph{priority search} \citep{arya1993algorithms,Silpa2008} or \emph{multi-probe LSH} \citep{Lv2007}, depending on whether the index structure is tree-based or hashing-based, respectively. For clarity, we do not consider backtracking---that is mainly a memory-saving technique---in the analysis below: it can be easily incorporated both into lookup search and our method by allowing the structure function $q$ to return a set of indices instead of a single index, and considering $\bigcup_{l \in q(x)} R_l$ instead of the single partition element $R_{q(x)}$.}: they select the corpus point into the candidate set if and only if it belongs to the same partition element as the query point. Consequently, their objective is to learn a partition in which the $k$ nearest neighbors of a query point belong to the same partition element with it. In contrast, our objective is to directly learn a partitioning classifier that predicts its nearest neighbors correctly. We will elucidate the difference in the next section.

\section{Candidate set selection for ANN search}
\label{sec:candidate_set_selection}

In view of the multilabel formulation of Sec. \ref{sec:ann2multilabel}, the natural classifier defined in Sec. \ref{sec:partitioning_classifiers} can directly be used to select a candidate set for ANN search. However, in this section we interpret also the earlier lookup-based candidate set selection methods (lookup search and voting search) in our multilabel classification framework, and show that they define a classifier of a different form. We show that this classifier---that we call the \emph{naive classifier}---is in fact the natural classifier for the different multilabel classification problem where the labels do not represent the corpus points that are the nearest neighbors of the query point but the corpus points that belong to the same partition element as the query point. This suggests that the natural classifier would be a more suitable candidate set selection method for ANN search than the naive classifier. The empirical results of Sec. \ref{sec:experiments} indicate that this is indeed the case.

% The natural classifier is a plug-in classifier that thresholds the maximum likelihood estimates (under the piecewise constant model and the widely used PAL and OVA multilabel problem reductions) for the conditional label probabilities (see Appendix \ref{sec:data_structures}), and the Bayes classifier is also obtained by thresholding these conditional label probabilities.

% In this section, we describe how the natural classifier described in Sec. \ref{sec:partitioning_classifiers} can be used the select the candidate set for ANN search, and how the earlier candidate set selection methods (lookup search and voting) define a classifier of a different form---that we call a naive classifier.

\paragraph{Candidate set selection for a single partition.} First, assume that we utilize the single fixed partition $\mathcal{P} = \{R_1, \dots, R_L\}$ and the training set $\{(x_i, y_i)\}_{i=1}^n$ to approximate the nearest neighbors of the query point $x$. The natural classifier defined in Sec. \ref{sec:partitioning_classifiers} selects the candidate set as
\begin{equation}
\label{eq:natural_classifier}
\begin{split}
\hat{L}(x) &= \{j \in \{1, \dots, m\} \,|\, \hat{\eta}_j(x) > \tau\},
\end{split}
\end{equation}
where $\tau \in [0,1]$, and the conditional label probability estimates $\hat{\eta}_j(x) = \frac{1}{N_{q(x)}}\sum_{i \,: \, x_i \in R_{q(x)}} y_{ij}$ are obtained as the observed label proportions among the training set points that belong to the same partition element with the query point. In contrast, lookup search selects the candidate set as
\begin{equation}
\label{eq:naive_classifier}
\begin{split}
\hat{L}(x) &= \{j \in \{1, \dots, m\} \,|\,  c_j \in R_{q(x)}\}, \\
% &= \{j \in \{1, \dots, m\} \,|\,  \Tilde{\eta}_j(x) > \tau\},
\end{split}
\end{equation}
i.e., it selects the corpus point into the candidate set if and only if it belongs to the same partition element with the query point. When interpreted in the classification framework of Sec. \ref{sec:ann2multilabel}, \eqref{eq:naive_classifier} defines the classifier $\hat{L}(x) = \{j \in \{1, \dots, m\} \,|\,  \Tilde{\eta}_j(x) > \tau\}$, where  $\tau \in [0,1)$ and $\Tilde{\eta}_j(x) = \mathbbm{1}_{R_{q(x)}}(c_j) 
%= \frac{1}{N_{q(x)}}\sum_{i \,:\, x_i \in R_{q(x)}} \mathbbm{1}_{R_{q(x_i)}}(c_j)
$; we call this a \emph{naive classifier}.

We immediately observe that the naive classifier is not a natural classifier for the multilabel classification problem in which the %training set $\{(x_i, y_i)\}_{i=1}^n$, where
labels are defined as $y_{ij} = \mathbbm{1}_{\nn_k(x_i)}(c_j)$
%for each $j=1, \dots, m$, is defined 
as in Sec.~\ref{sec:ann2multilabel}. Instead, it is a natural classifier for the different multilabel classification problem in which the %training set is $\{(x_i,\Tilde{y}_{i})\}_{i=1}^n$, where
labels are defined as $\Tilde{y}_{ij} = \mathbbm{1}_{R_q(x_i)}(c_j)$ %for each $j = 1, \dots, m$
. In other words, the naive classifier is geared towards the learning problem in which---instead of the $k$ nearest neighbors of the query point---the labels represent the corpus points that belong to the same partition element as the query point. The candidate set selection method \eqref{eq:naive_classifier} also explains why the objective of the earlier supervised methods for \emph{learning} the partition differs from ours: in these methods, the objective is to maximise the number of nearest neighbors of the query point that belong to the same partition element with it in order to maximise the recall (while minimising the number of non-neighbors in that element in order to maximise precision).

\paragraph{Candidate set selection for an ensemble of partitions.} Assume that the fixed set of partitions $\{\mathcal{P}_{t}\}_{t=1}^T$ is used to approximate the $k$ nearest neighbors of a query point $x$. The natural classifier defined in Sec. \ref{sec:partitioning_classifiers} selects the candidate set 
\begin{equation}
\label{eq:natural_classifier_ensemble}
\begin{split}
\hat{L}(x) &= \{j \in \{1, \dots, m\} \,|\, \hat{\eta}_j(x) > \tau\},
\end{split}
\end{equation}
as in \eqref{eq:natural_classifier}, but now $\hat{\eta}_j(x) =  \frac{1}{T}\sum_{t=1}^T \hat{\eta}^{(t)}_j(x)$, where the contributions of the individual partitions $\hat{\eta}^{(t)}_j(x)$ are defined as above. In contrast, the earlier (both supervised and unsupervised) methods select the corpus point into the candidate set if and only if it belongs to the same partition element as the query point in at least one of the $T$ partitions. Hence, the candidate set selected by lookup search is
\begin{equation}
\label{eq:naive_classifier_ensemble}
\begin{split}
\hat{L}(x) &= \left\{j \in \{1, \dots, m\} \,|\,  c_j \in \bigcup_{t=1}^T R^{(t)}_{q^{(t)}(x)}\right\} = \{j \in \{1, \dots, m\} \,|\,  \Tilde{\eta}_j(x) > \tau\},
\end{split}
\end{equation}
where $\Tilde{\eta}_j(x) := \frac{1}{T} \sum_{t=1}^T \Tilde{\eta}_j^{(t)}(x)$, $\tau \in [0, \frac{1}{T})$, and the contributions of the partitions $\Tilde{\eta}_j^{(t)}(x)$ are defined as above. 

Unlike in the case of a single partition---where the value of the threshold parameter $\tau \in [0,1)$ does not affect the classification decision of the naive classifier, since $\Tilde{\eta}_j(x) \in \{0,1\}$---now $\tau$ affects the classification decision, since $\Tilde{\eta}_j(x) \in \{0, \frac{1}{T}, \dots, \frac{T-1}{T}, 1\}$. Hence, a tuning parameter can be added to lookup search by allowing $\tau$ to be chosen freely as proposed by~\citet{Hyvonen2016} who call the resulting method \emph{voting search}.

\section{Consistency of partitioning classifiers for ANN search}
\label{sec:theory}

The ideal index structure for ANN search always returns a candidate set that contains all the $k$ nearest neighbors of the query point and no other corpus points. Under the multilabel formulation, this corresponds to a classifier for which the expected multilabel 0-1 loss $EL(g(X),Y) = P\{g(X) \neq Y\}$ is zero. To this end, we prove a sufficient condition for the consistency of a partitioning classifier for ANN search under 0-1 loss. Consistency under 0-1 loss also directly implies consistency for the other common multilabel loss functions, such as Hamming loss, precision, recall, and $F$-measures. As a concrete example, we prove the consistency of the chronological $k$-d tree \citep{Bentley1975} by checking that this condition holds for it.

\subsection{Sufficient condition for consistency}
\label{sec:theory_sufficient_condition}

The classical theorem for proving consistency of partitioning classifiers for binary classification is:

\begin{theorem} (\cite{devroye2013probabilistic}, Theorem 6.1, p. 94--95)
\label{theorem:general_consistency}
Assume that only the features $X_1, \dots, X_n$ are used to learn the partition $\mathcal{P}^{(n)} = \pi(X_1, \dots, X_n)$. The natural classifier\footnote{The natural classifier for binary classification is defined as the classifier that classifies the query point into the majority class of the training set points that belong to the same partition element with it.} $g^{(n)}$ defined by $\mathcal{P}^{(n)}$  is consistent (under 0-1 loss) for binary classification, if 
\begin{enumerate}[(i)]
\item $N_{q(X)} \rightarrow \infty$ in probability, and
\item $\diam\left(R_{q(X)}\right) \rightarrow 0$ in probability,
\end{enumerate}
when $n \rightarrow \infty$.
\end{theorem}
The number of the training set points in the partition element the query point $x$ belongs to is denoted by $N_q(x) := |\{i \,:\, X_i \in R_{q(x)}\}|$, and the diameter of a set $A$ is defined as the maximum distance between any two points of this set and denoted by $\mathrm{diam}(A) := \underset{a,\,b \,\in\, A}{\sup} \|a - b\|$. 

While this result is for binary classification, %in view of \eqref{eq:reduction2binary} 
it can be readily extended to the multilabel case. However, as a multilabel classification problem, ANN search has two distinguishing properties: $(i)$ the Bayes error $\mathcal{R}^*$ is zero; $(ii)$ decision boundaries between the labels consist of subsets of hyperplanes. It turns out that in this case, the second condition of Theorem \ref{theorem:general_consistency} is sufficient for the consistency of a partitioning classifier:

\begin{theorem}
\label{theorem:consistency_diameter}
Let $g^{(n)}$ be a natural classifier defined by the partition $\mathcal{P}^{(n)} = (R_1, \dots, R_L)$ and the threshold parameter $\tau \in [0,1)$ for ANN search. Assume that the distribution of $X$, denoted by $\mu$, is continuous. If $\mathrm{diam}(R_{q(X)}) \rightarrow 0$ in probability---that is, if for every $\epsilon > 0$,
\[
P \{\mathrm{diam}(R_{q(X)}) > \epsilon\} \rightarrow 0
\] when $n \rightarrow \infty$, then the classifier $g^{(n)}$ is consistent (for 0-1 loss)---i.e., $E_{D_n}\mathcal{R}(g^{(n)}) \rightarrow 0$.
\end{theorem}

\begin{proof}
If for all the pairs of corpus points $(c_j, c_{j^\prime})$, $j^\prime \neq j$, all the points of the partition element $R_l$ are closer to $c_j$ than $c_{j^\prime}$ (or vice versa)---that is, if there is no such pair $(c_j, c_{j^\prime})$ for which there exists  $a,b \in R_l$ such that $\|a-c_j\| <  \|a - c_{j^\prime}\|$ and $\|b-c_j\| >  \|b - c_{j^\prime}\|$---then also $\hat{\eta}_j(x) = \eta_j(x)$ for each $x \in R_l$ and $j = 1, \dots, m$; consequently, each $x \in R_l$ is classified correctly for any $\tau \in [0,1)$. Now, since for each $j = 1, \dots, m$, 
\[
\begin{split}
&\,\,P\{g^{(n)}_j(X) \neq \eta_j(X)\} \\ 
\leq &\,\,P\left( \exists j^\prime \neq j : \exists a, b \in R_{q(X)} \,\,\mathrm{s.t.}\,\,  \|a-c_j\| <  \|a - c_{j^\prime}\|, \, \|b-c_j\| >  \|b - c_{j^\prime}\| \right) \\
\leq &\,\,\sum_{j^\prime \neq j} P \{\exists a, b \in R_{q(X)} \,\,\mathrm{s.t.}\,\, \|a-c_j\| <  \|a - c_{j^\prime}\|, \, \|b-c_j\| >  \|b - c_{j^\prime}\|\},
\end{split}
\]
to prove consistency of $g^{(n)}$ it is sufficient to show that for all $j, j^\prime \in \{1, \dots, m\}$, $j \neq j^\prime$,
\[
P\{\exists a, b \in R_{q(X)} \,\,\mathrm{s.t.}\,\,  \|a-c_j\| <  \|a - c_{j^\prime}\|, \, \|b-c_j\| >  \|b - c_{j^\prime}\|\} \rightarrow 0
\]
in probability when $n \rightarrow \infty$.

Choose any $j, j^\prime$, $j \neq j^\prime$, and denote the hyperplane that is halfway in between the corpus points $c_j$ and $c_{j^\prime}$ by $H : = \{x \in \mathbb{R}^d \,:\, \|x-c_j\| =  \|x - c_{j^\prime}\|\}$. For any $t = 1,2, \dots$, let $H_t$ denote the set surrounding $H$ by a margin of width $1 / t$. Since $H_1 \supset H_2 \supset H_3 \dots, $ and $H = \cap_{t=1}^\infty H_t$, it follows from the upper continuity of the probability measure that $\underset{t\rightarrow\infty}{\lim} \mu(H_t) = \mu(H)$. Because the Lebesgue measure of the hyperplane $H$ in $\mathbb{R}^d$ is zero and $\mu$ is absolutely continuous w.r.t. the Lebesgue measure by the assumption, then also $\underset{t\rightarrow\infty}{\lim} \mu(H_t)  = \mu(H) = 0$.

Now, for any $t = 1,2, \dots$, if $R_{q(x)}$ crosses the hyperplane $H$, then either $x \in H_t$ or the diameter of the $R_{q(x)}$ is greater than $1/t$. Hence, 
\[
\begin{split}
&\,\, P\{\exists a, b \in R_{q(X)} \,\,\mathrm{s.t.}\,\,  \|a-c_j\| <  \|a - c_{j^\prime}\|, \, \|b-c_j\| >  \|b - c_{j^\prime}\|\}  \\
\leq &\,\,P\{X \in H_t \,\, \mathrm{or} \,\, \mathrm{diam}(R_{q(X)}) > 1/t\}  \\
\leq &\,\,\mu(H_t) + P\{\mathrm{diam}(R_{q(X)} ) > 1/t\}.  
\end{split}
\]
We can get $\mu(H_t)$ as small as desired by choosing a large enough $t$; and since by assumption the second term is arbitrarily small when $n$ is large enough, the result follows.
\end{proof}

\subsection{Consistency of chronological k-d tree}
\label{sec:theory_kd_tree}

Next, we illustrate the utility of Theorem \ref{theorem:consistency_diameter} by applying it to prove the consistency of the \emph{chronological $k$-d tree} \citep{Bentley1975} that rotates the split directions and uses the same split direction for all the nodes at one level of a tree. At the first level the training data is split at the median of the first coordinates of the data points. At the second level both nodes are split at the median of the second coordinates of the node points. At the $(d+1)$th level, the nodes are split again at the median of the first coordinates, and so on (see Appendix \ref{sec:chronological_kd_tree}). 

More precisely, let $X, X_1, \dots, X_n \in \mathbb{R}^d$ be i.i.d. random variables. A chronological $k$-d tree can be formalized as a partitioning rule $\pi$ that returns the partition $\mathcal{P}^{(n)} = \pi(X_1, \dots, X_n)$. When the tree height is $\ell$, this partition has $2^\ell$ elements (also called \emph{leafs}). The leafs are hyperrectangles in $\mathbb{R}^d$. Some of the edges of these hyperrectangles may have an infinite length.
To handle %the leafs with an infinite length
these leafs, we introduce the notation where, for any $M > 0$, the hypercube $[-M,M]^d$ divides the partition elements $R_1, \dots, R_{2^\ell}$ into three disjoint sets:
\begin{equation}
\label{eq:decomposition}
\begin{split}
A &:= \{l \in \{1, \dots, 2^\ell\} \,:\, R_l \subset [-M,M]^d\}, \\
C &:= \{l \in \{1, \dots, 2^\ell\} \,:\, R_l \subset \mathbb{R}^d \setminus [-M,M]^d\}, \\
B &:= \{1, \dots, 2^\ell\} \setminus (A \cup C).
\end{split}
\end{equation}
Here $A$ is the set of indexes of the partition elements that are completely inside the hypercube $[-M,M]^d$, $B$ is the set of indexes of the partition elements that cross its boundary, and $C$ is the set of indexes of the partition elements that are completely outside of it. 

First, we prove two auxiliary results that bound the number of nodes crossing the boundary of the box $[-M,M]^d$ and the combined length of the edges (in any fixed coordinate direction) of the nodes that reside completely inside $[-M,M]^d$, respectively. Note that these bounds are of purely combinatorial nature and thus do not depend on the training set. The proofs of the following results are presented in Appendix \ref{sec:proofs}.

\begin{lemma}
\label{lemma:node_sum}
For any training set $D_n$, it holds for the number of nodes of a chronological $k$-d tree---denoted by $N_B := |B|$---crossing the border of the hypercube $[-M,M]^d$ that
\[
N_B  \leq 4d \cdot 2^{\ell - \frac{\ell}{d}}.
\]
\end{lemma}

\begin{lemma}
\label{lemma:edge_bound}
Let $j \in \{1, \dots, d\}$ be any coordinate direction. Denote the length of the node $R_l$ in the $j$th coordinate direction by $V_l$. Then for any training set $D_n$, 
\[
\sum_{l \in A} V_l \leq  4M \cdot 2^{\ell - \frac{\ell}{d}}.
\]
\end{lemma}

We are now in a position to establish the consistency of the chronological $k$-d tree for approximate nearest neighbor search. In view of Theorem \ref{theorem:consistency_diameter} it suffices to prove that the leaf diameter converges to zero in probability:

\begin{theorem} 
If for the height of a chronological $k$-d tree holds that $\ell \rightarrow \infty$ when $n \rightarrow \infty$, then the leaf diameter $\diam (R_{q(X)})$ converges to zero in probability.
\end{theorem}

\section{Experiments}
\label{sec:experiments}

We present empirical results validating the utility of our framework. In particular, we compare the natural classifier to the earlier candidate set selection methods discussed in Sec. \ref{sec:candidate_set_selection} for different types of unsupervised trees that have been widely used for ANN search. Specifically, we use ensembles of randomized $k$-d trees~\citep{Friedman1976,Silpa2008}, random projection (RP) trees~\citep{Dasgupta2008,Hyvonen2016}, and principal component (PCA) trees~\citep{Sproull1991,Jaasaari2019} (see Appendix \ref{sec:data_structures} for detailed descriptions of these data structures). Another consequence of the multilabel formulation of Sec. \ref{sec:ann2multilabel} is that it enables using any established multilabel classifier for ANN search. To demonstrate this concretely, we train a random forest consisting of standard multilabel classification trees (trained under the PAL reduction \citep{reddi2019stochastic} by using multinomial log-likelihood as a split criterion) and use it as an index structure for ANN search; it turns out that the fully supervised classification trees have an improved performance compared to the earlier unsupervised trees on some---but, curiously, not on all---data sets.

We follow a standard ANN search performance evaluation setting \citep{Aumuller2019a,Li2019} by using the corpus as the training set, searching for $k=10$ nearest neighbors in Euclidean distance, and measuring performance by evaluating average recall and query time over the test set of 1000 points. We use four benchmark data sets: Fashion ($m = 60000$, $d = 784$), GIST ($m = 1000000$, $d = 960$), Trevi ($m = 101120$, $d = 4096$), and STL-10 ($m = 100000$, $d = 9216$). All the algorithms are implemented in C++ and run using a single thread. We tune the hyperparameters by grid search and plot the Pareto frontiers of the optimal hyperparameters. Further details of the experimental setup are found in Appendix \ref{sec:experimental_setup}. The code used to produce the experimental results is attached as supplementary material and can also be found at \url{https://github.com/vioshyvo/a-multilabel-classification-framework}.

\begin{figure*}[!hbtp]
\centering
%\vspace*{-4mm}
\includegraphics[width=\textwidth]{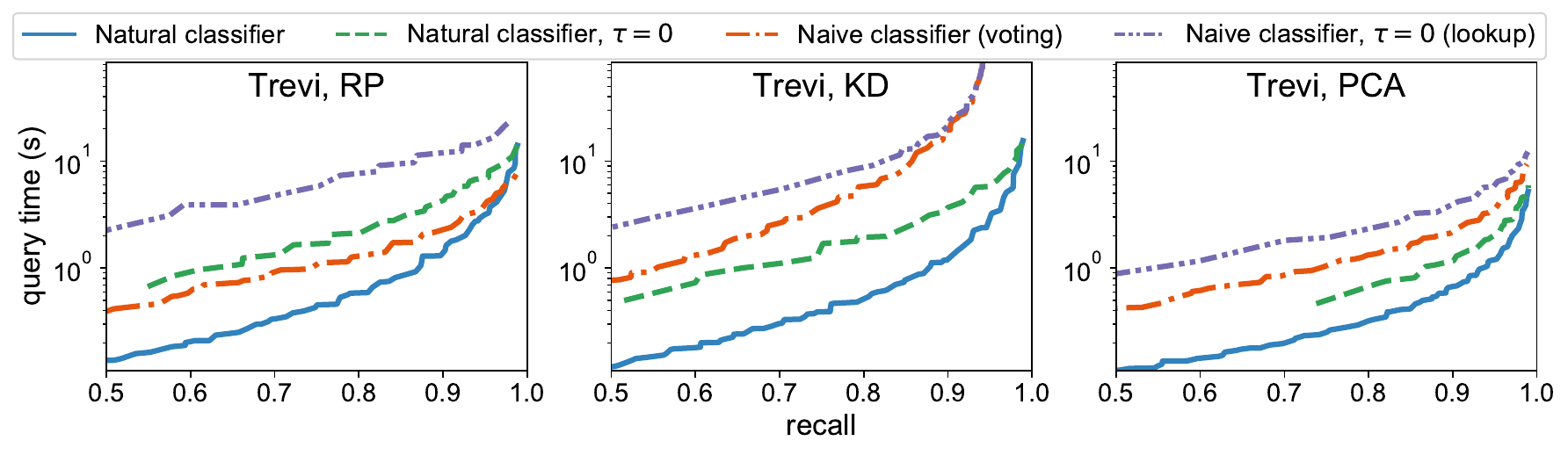}%\vspace*{-6mm}
\caption{Recall vs. query time (log scale) of ensembles of, RP, $k$-d, and PCA trees. The solid blue line is the natural classifier proposed in this paper; the dash-dotted red line is the natural classifier with $\tau = 0$ that is included for completeness; the dashed green line is voting; and the double-dash-dotted violet line is lookup search. The natural classifier is the fastest and the lookup search is the slowest of the methods for each tree type.}
\label{fig:search_methods}
\end{figure*}

\paragraph{Comparison of candidate set selection methods.} The candidate set selection method proposed in this article is the natural classifier \eqref{eq:natural_classifier_ensemble} described in Sec. \ref{sec:partitioning_classifiers}; for completeness, we also include the special case obtained by fixing $\tau = 0$ in the comparison. The earlier methods are lookup search (naive classifier \eqref{eq:naive_classifier_ensemble} with $\tau=0$) and voting \citep{Hyvonen2016,Jaasaari2019} (naive classifier \eqref{eq:naive_classifier_ensemble} with $\tau$ as a free tuning parameter). The results for the Trevi data set are presented in Fig. \ref{fig:search_methods} and indicate, as the discussion of Sec. \ref{sec:candidate_set_selection} suggests, that the natural classifier performs better than the earlier lookup-based methods for all types of trees (this finding holds consistently over all the data sets in our experiments; see Fig. \ref{fig:search_methods_full} in Appendix).

% the performance of partition-based index structures can indeed be improved by interpreting them as partitioning classifiers, and, consequently, selecting the candidate set by thresholding the probability estimates induced by them (this finding holds consistently over all the data sets in our experiments; see Appendix \ref{sec:additional_experimental_results}). 

% The results for the Trevi data set are presented in Fig. \ref{fig:search_methods} and indicate that, as the discussion of Sec. \ref{sec:related_work} suggests, the performance of partition-based index structures can indeed be improved by interpreting them as partitioning classifiers, and, consequently, selecting the candidate set by thresholding the probability estimates induced by them (this finding holds consistently over all the data sets in our experiments; see Appendix \ref{sec:additional_experimental_results}). 

\paragraph{Comparison of tree types. } We compare the aforementioned ensembles of unsupervised (RP, KD, and PCA) trees and the random forest consisting of supervised classification trees (RF); for all four tree types the candidate set is selected by \eqref{eq:natural_classifier_ensemble}. The results are shown in Table \ref{table:comparison}. Since the random forest (RF) leverages supervised information to learn the trees, we would expect that it is the fastest tree-based method. Indeed, this is the case on Fashion and GIST. However, on STL-10 and Trevi, the unsupervised PCA tree is the fastest method. We hypothesize that this is because of the high dimensionality of STL-10 and Trevi: standard supervised classification trees employed by random forest are restricted to axis-aligned splits, whereas PCA trees---although they use an unsupervised split criterion---can find more informative oblique split directions. An interesting topic for future work would be to apply supervised classification trees that can utilize oblique split directions.

\begin{table*}[hbtp]
\begin{center}
\caption{Query times (seconds / 1000 queries) at different recall levels for the different tree types. The fastest method in each case is typeset in boldface. }
\label{table:comparison}
\begin{tabular}{l c c c c c}
\toprule
data set & R (\%) & PCA & KD & RP & RF \\
\midrule
 & 80 & 0.075 & 0.076 & 0.099 & {\bf 0.063} \\
Fashion & 90 & 0.111 & 0.126 & 0.172 & \bf{0.095}  \\
 & 95 & 0.163 & 0.171 & 0.261 & \bf{0.146} \\
\midrule
 & 80 & 1.330 & 0.958 & 1.009 & \bf{0.705} \\
GIST & 90 & 2.942 & 2.286 & 2.226 & \bf{1.530}  \\
 & 95 & 5.641 & 4.451 & 4.598 & \bf{3.253}  \\
\midrule
 & 80 & {\bf 0.382} & 0.872 & 1.211 & 0.756  \\
STL-10 & 90 & {\bf 0.756} & 2.126 & 3.248 & 1.774  \\
 & 95 & {\bf 1.315} & 4.376 & 7.330 & 3.654   \\
\midrule
 & 80 & {\bf 0.330} & 0.543 & 0.591 & 0.582 \\
Trevi & 90 & {\bf 0.684} & 1.464 & 1.468 & 1.234  \\
 & 95 & {\bf 1.212} & 3.244 & 3.289 & 2.350  \\
\bottomrule
\end{tabular}
\end{center}
\end{table*}

\section{Conclusion}
\label{sec:conclusion}

We establish a general theoretical framework for ANN search by formulating candidate set selection as a multilabel learning task. Empirical results validate our framework: a natural classifier derived directly from the problem formulation is a strict improvement over the earlier lookup-based candidate set selection methods. In addition, we provide a sufficient condition that guarantees consistency of a partitioning classifier for ANN search. We verify this condition for chronological $k$-d trees, indicating that---given enough training data---they retrieve a candidate set containing all the $k$ nearest neighbors of the query point and no other corpus points. 

\paragraph{Limitations. } Supervised ANN search methods typically have longer pre-processing times compared to unsupervised methods. This is because (1) they require computing the true nearest neighbors $\{\ve{y}_i\}_{i=1}^n$ of the training set points $\{\ve{x}_i\}_{i=1}^n$ and (2) supervised index structures are often slower to build compared to their unsupervised counterparts (c.f. Appendix \ref{sec:index_construction_times}). If fast index construction is required, the second problem can be mitigated by learning trees in an unsupervised fashion, but using them as partitioning classifiers as described in Sec. \ref{sec:partitioning_classifiers}, since the experiments of Sec. \ref{sec:experiments} suggest that the candidate set selection method has a more pronounced effect on the performance than the tree type.

\paragraph{Future research directions.} While we demonstrate our approach using a random forest classifier, we expect that the most important consequence of our work is that it enables using any type of classifier as an index structure for ANN search. In particular, gradient boosted trees~\citep{Friedman2001} are promising since they are often more accurate than random forests. \emph{Extreme classification} models, including tree-based models~\citep{Agrawal2013,Prabhu2014,Jain2016}, sparse linear models~\citep{Babbar2017,Babbar2019,Yen2017}, and embedding-based neural networks~\citep{Guo2019}, are also promising model candidates for ANN search since they are specifically tailored for multilabel classification problems with extremely large label spaces. 

Our formulation enables analyzing ANN search in the statistical learning framework, thus opening multiple theoretical research questions: (1) Can we establish a sufficient condition for \emph{strong} consistency? (2) Can we prove consistency of more adaptive partitioning classifiers, such as PCA trees or classification trees? (3) Can we establish faster than logarithmic convergence rates? The last question
is especially interesting, since prediction times of trees are logarithmic: a positive answer would
theoretically justify decreasing query times by increasing the training set size. 

\section*{Acknowledgements}

Funding in direct support of this work: Academy of Finland grants \#345635 (DAISY), \#311277 (TensorML), and \#313857 (WiFiUS). The authors wish to thank the Finnish Computing Competence Infrastructure (FCCI) for supporting this project with computational and data storage resources.

%Use unnumbered first level headings for the acknowledgments. All acknowledgments go at the end of the  aper before the list of references. Moreover, you are required to declare funding (financial activities supporting the submitted work) and com eting interests (related financial activities outside the submitted work). More information about this disclosure can be found at: \url{https://neurips.cc/Conferences/20%22/PaperInformation/FundingDisclosure}.

%Do {\bf not} include this section in the anonymize submission, only in the final paper. You can use the \texttt{ack} environment provided in the style file to autmoatically hide this section in the anonymized submission.

\bibliography{neurips_2022}

%%%%%%%%%%%%%%%%%%%%%%%%%%%%%%%%%%%%%%%%%%%%%%%%%%%%%%%%%%%%
\section*{Checklist}

%%% BEGIN INSTRUCTIONS %%%
% The checklist follows the references.  Please
% read the checklist guidelines carefully for information on how to answer these
% questions.  For each question, change the default \answerTODO{} to \answerYes{},
% \answerNo{}, or \answerNA{}.  You are strongly encouraged to include a {\bf
% justification to your answer}, either by referencing the appropriate section of
% your paper or providing a brief inline description.  For example:
% \begin{itemize}
%   \item Did you include the license to the code and datasets? \answerYes{See Section~\ref{gen_inst}.}
%   \item Did you include the license to the code and datasets? \answerNo{The code and the data are proprietary.}
%   \item Did you include the license to the code and datasets? \answerNA{}
% \end{itemize}
% Please do not modify the questions and only use the provided macros for your
% answers.  Note that the Checklist section does not count towards the page
% limit.  In your paper, please delete this instructions block and only keep the
% Checklist section heading above along with the questions/answers below.
% %%% END INSTRUCTIONS %%%

\begin{enumerate}

\item For all authors...
\begin{enumerate}
  \item Do the main claims made in the abstract and introduction accurately reflect the paper's contributions and scope?
    \answerYes{}
  \item Did you describe the limitations of your work?
    \answerYes{} See Section \ref{sec:conclusion}
  \item Did you discuss any potential negative societal impacts of your work?
    \answerNA{}
  \item Have you read the ethics review guidelines and ensured that your paper conforms to them?
    \answerYes{}
\end{enumerate}

\item If you are including theoretical results...
\begin{enumerate}
  \item Did you state the full set of assumptions of all theoretical results?
    \answerYes{} 
        \item Did you include complete proofs of all theoretical results?
    \answerYes{} See Appendix~\ref{sec:proofs}
\end{enumerate}

\item If you ran experiments...
\begin{enumerate}
  \item Did you include the code, data, and instructions needed to reproduce the main experimental results (either in the supplemental material or as a URL)?
    \answerYes{} See the supplemental repository or \url{https://github.com/vioshyvo/a-multilabel-classification-framework}
  \item Did you specify all the training details (e.g., data splits, hyperparameters, how they were chosen)?
    \answerYes{} See Appendix  \ref{sec:experimental_setup} and the supplemental repository or \url{https://github.com/vioshyvo/a-multilabel-classification-framework}
        \item Did you report error bars (e.g., with respect to the random seed after running experiments multiple times)?
    \answerNo{}
        \item Did you include the total amount of compute and the type of resources used (e.g., type of GPUs, internal cluster, or cloud provider)?
    \answerNo{} 
\end{enumerate}

\item If you are using existing assets (e.g., code, data, models) or curating/releasing new assets...
\begin{enumerate}
  \item If your work uses existing assets, did you cite the creators?
    \answerNA{}
  \item Did you mention the license of the assets?
    \answerNA{}{}
  \item Did you include any new assets either in the supplemental material or as a URL?
    \answerNA{}
  \item Did you discuss whether and how consent was obtained from people whose data you're using/curating?
    \answerNA{}
  \item Did you discuss whether the data you are using/curating contains personally identifiable information or offensive content?
    \answerNA{}
\end{enumerate}

\item If you used crowdsourcing or conducted research with human subjects...
\begin{enumerate}
  \item Did you include the full text of instructions given to participants and screenshots, if applicable?
    \answerNA{}
  \item Did you describe any potential participant risks, with links to Institutional Review Board (IRB) approvals, if applicable?
    \answerNA{}
  \item Did you include the estimated hourly wage paid to participants and the total amount spent on participant compensation?
    \answerNA{}
\end{enumerate}

\end{enumerate}

%%%%%%%%%%%%%%%%%%%%%%%%%%%%%%%%%%%%%%%%%%%%%%%%%%%%%%%%%%%%

\newpage

\appendix

\section{Proofs}
\label{sec:proofs}

% \subsection*{Proof of Theorem 2}

\subsection*{Proof of Lemma 1}

\begin{proof}
The border of the hypercube consists of $(d-1)$-dimensional faces.  Choose a coordinate direction $j \in \{1, \dots, d\}$ and consider a $(d-1)$-face that is orthogonal to that coordinate axis. Denote the number of nodes crossing that $(d-1)$-face by $N_B^{(j)}$. Before any splits there is one node---the whole feature space $\mathbb{R}^d$---that crosses it. Splitting a node that crosses that $(d-1)$-face at a coordinate direction other than $j$ creates two nodes crossing it if the splitting hyperplane intersects with $[-M,M]^d$ (if the splitting hyperplane does not intersect with $[-M,M]^d$ then it does affect $N_B^{(j)}$). Splitting at the $j$th direction does not increase $N_B^{(j)}$ since the splitting hyperplane is perpendicular to the $(d-1)$-face we consider and so cannot cross it.

Therefore, if $\ell$ is a multiple of $d$, we have $N_B^{(j)} \leq 2^{\ell - \frac{\ell}{d}}$ since each full round of $d$ splits contains $d-1$ splits orthogonal to the $j$th coordinate direction, each of which may double the number of nodes crossing the $(d-1)$-face, and one split parallel to the $j$th coordinate direction that doesn't increase the number. If $\ell$ is not a multiple of $d$, then
\[
N_B^{(j)} \leq 2^{\ell - \lfloor\frac{\ell}{d} \rfloor} \leq 2^{\ell - \frac{\ell}{d} + 1}
\]
because the last incomplete round of splits may not contain a split at the $j$th coordinate direction. Since for each coordinate direction a $d$-dimensional hypercube has two $(d-1)$-faces that are orthogonal to that coordinate axis, we have\footnote{Since we are proving an upper bound it does not matter that we double count nodes that cross more than one $(d-1)$-face.}
\[
N_B \leq 2\sum_{j=1}^d N_B^{(j)} \leq  4d \cdot 2^{\ell - \frac{\ell}{d}}.
\]
\end{proof}

\subsection*{Proof of Lemma 2}

\begin{proof}[\unskip\nopunct]
For each $l = 1, \dots, 2^\ell$, denote the length of the hyperrectangle $R_l^\prime := R_l \cap [-M,M]^d$ in the $j$th coordinate direction by $V_l^\prime$. Clearly, $V_l = V_l^\prime$  for each $l \in A$ by the definition of the set $A$. Thus, 
\[
\sum_{l \in A} V_l = \sum_{l \in A} V_l^\prime \leq \sum_{l=1}^{2^\ell} V_l^\prime.
\]

Before any splits, there is one node with $V_l^\prime = 2M$.
Splitting a node in a coordinate direction other than $j$ creates two child nodes with the same length in the $j$th coordinate direction as the parent node, and thus doubles the contribution of the parent node to the sum over the nodes\footnote{Here we assume that the splitting hyperplane intersects with $[-M,M]^d$. If the splitting hyperplane does not intersect with $[-M,M]^d$, then the split does not increase $\sum_{l=1}^{2^\ell} V_l^\prime$. Thus, the inequality in \eqref{eq:node_sum_even} holds as an equality if and only if all the splitting hyperplanes that are not orthogonal to the $j$th coordinate direction intersect with $[-M,M]^d$.}. When we split a node in the $j$th coordinate direction, the \emph{sum} of the lengths of the child nodes in the $j$th direction equals the length of the parent node in that direction; thus, the split does not affect the sum over the nodes. Hence, we have
\begin{equation}
\label{eq:node_sum_even}
\sum_{l=1}^{2^\ell} V_l^\prime \leq 2M \cdot 2^{\ell - \frac{\ell}{d}}
\end{equation}
when $\ell$ is a multiple of $d$. When $\ell$ is not a multiple of $d$, the last incomplete round may not contain a split in the $j$th coordinate direction, and thus
\[
\sum_{l=1}^{2^\ell} V_l^\prime \leq 2M \cdot 2^{\ell - \lfloor \frac{\ell}{d} \rfloor} \leq 4M \cdot 2^{\ell - \frac{\ell}{d}}.
\]

\end{proof}

% \subsection*{Lemma 3}

For completeness, we prove here the following well-known variation of Markov's inequality.

\begin{lemma}
\label{lemma:markov}
Suppose $H$ is an event, $a > 0$, and $X$ is a non-negative random variable for which $E|X| < \infty$. Then,
\[
P\{X > a, H\} \leq \frac{E[\mathbbm{1}_H X]}{a}.
\]
\end{lemma}

\begin{proof}
We can write
\[
E[\mathbbm{1}_H X] \geq E[\mathbbm{1}_H X \mathbbm{1}\{X > a\}] \geq a E[  \mathbbm{1}_H \mathbbm{1}\{X > a\}] = a P\{X > a, H\}, \\
\]
and the result follows by dividing by $a$.
\end{proof}

\subsection*{Proof of Theorem 3}

\begin{proof}
Choose a coordinate direction $ j \in \{1, \dots, d\}$ and denote the length of an edge of the hyperrectangle $R_l$ in that direction by $V_l$. Since the coordinate direction was chosen arbitrarily, it suffices to show that $V_{q(X)}$ converges to zero in probability to prove that also the cell diameter converges to zero in probability.

For any training set size $n$, define $\ell^\prime := \min (\ell, \lfloor \log_2 \log_2 n \rfloor)$. Since for any training set\footnote{To keep notation consistent throughout the article, we denote the training set also here by $D_n:= \{(X_i, Y_i)\}_{i=1}^n$. However, it should be observed that the chronological $k$-d tree uses only the inputs $X_1, \dots, X_n$ to learn the partition; thus, the learned partition does not depend on the labels $Y_1, \dots, Y_n$.} $D_n$ the probability $P\{V_{q(X)} > \delta \,|\, D_n\}$ is non-increasing w.r.t. to the tree height and $\ell^\prime \leq \ell$, in order to prove that $P\{V_{q(X)} > \delta\} \rightarrow 0$ for the original tree height $\ell$, it is sufficient to show that it goes to zero for the tree height $\ell^\prime$. 

For any $\theta >0 $, there exists $M>0$ s.t. $P\{X \notin [-M,M]^d\} < \theta$. The box $[-M,M]^d$ divides any partition of $\mathbb{R}^d$ into three disjoint sets $A$, $B$, and $C$ as defined in \eqref{eq:decomposition} that correspond to the indexes of the nodes completely inside $[-M,M]^d$, the indexes of the nodes crossing its border and the indexes of the nodes completely outside of it, respectively\footnote{We define the sets $A$, $B$, and $C$ here for a tree of height $\ell^\prime$.}. We can now decompose the probability of the event $\{V_{q(X)} > \delta\}$ into three parts corresponding to the sets $A$, $B$, and $C$: 
\begin{equation}
\label{eq:partition1}
\begin{split}
P\{V_{q(X)} > \delta\}  &\leq P\{V_{q(X)} > \delta, q(X) \in A \} + P\{q(X) \in B\} + P\{q(X) \in C\}.
\end{split}
\end{equation}
 
Choose $\epsilon > 0$ and denote the event that no partition element has a probability mass larger than $(1 + \epsilon) / 2^{\ell^\prime}$ by
\[
G := \bigcap_{l=1}^{2^{\ell^\prime}} \left\{ \mu(R_l) \leq \frac{1 + \epsilon}{2^{\ell^\prime}}\right\},
\]
where $\mu(A) := P\{X \in A\}$ is the probability distribution of $X$ for any measurable set $A$. Our strategy is to first handle this case where all the leafs contain an approximately equal probability mass, and then bound the probability of $G^C$ by applying the Vapnik-Chervonenkis inequality to show the uniform convergence of the empirical distribution of $X$ to its true distribution in the class of leafs of a chronological $k$-d tree, i.e., in the class of hyperrectangles in $\mathbb{R}^d$. To this end, we further partition the right hand side of \eqref{eq:partition1} as  
\[
%\label{eq:first_eq2}
\begin{split}
P\{V_{q(X)} > \delta\}  &\leq 
\underbrace{P\{V_{q(X)} > \delta, q(X) \in A, G\}}_{I} 
+ \underbrace{P\{q(X) \in B, G\}}_{II} 
+ \underbrace{P(G^C)}_{III} 
+ \underbrace{P\{q(X) \in C\}}_{IV} %\\&=: (i) + (ii) + (iii) + (iv),
\end{split}
\]
and bound these four terms.
%We will bound each four terms on the right hand side of \eqref{eq:first_eq2} separately.

\emph{Term IV:} Since $P\{q(X) \in C \} \leq P\{X \notin [-M,M]^d\} < \theta$, we can get this term as small as desired by choosing a small enough $\theta$.

\emph{Term I:} By applying Lemma \ref{lemma:markov} (with the event $\{q(X) \in A\} \cap G$), we see that
\[
P\{V_{q(X)} > \delta, q(X) \in A, G\} \leq \frac{E \left[\mathbbm{1}_G V_{q(X)} \mathbbm{1}\{q(X) \in A\}  \right]}{\delta}.  
\]
Therefore, it suffices to bound $E\left[\mathbbm{1}_G V_{q(X)} \mathbbm{1}\{q(X) \in A\} \right]$, for which we have
\begin{equation}
\label{eq:second_term2}
\begin{split}
E\left[\mathbbm{1}_G V_{q(X)} \mathbbm{1}\{q(X) \in A\}  \right] &=E\left[\mathbbm{1}_G E\left[V_{q(X)} \mathbbm{1}\{q(X) \in A\}  \,|\, D_n\right]\right] \\
&= E\left[\mathbbm{1}_G\sum_{l =1 }^{2^{\ell^\prime}}\mu(R_l) V_l \mathbbm{1}\{l \in A\} \right] \\
&= E\left[\mathbbm{1}_G\sum_{l \in A} \mu(R_l) V_l \right] \\
&\leq \frac{1 + \epsilon}{2^{\ell^\prime}}  E \left[ \sum_{l \in A}  V_l \right]\\ 
&\leq \frac{1 + \epsilon}{2^{\ell^\prime}} \cdot  4M \cdot 2^{\ell^\prime -  \frac{\ell^\prime}{d}} \\
&= 4 M \cdot \frac{1 + \epsilon}{2^{\ell^\prime / d}},
\end{split}
\end{equation}  
where the outermost expectation on the right hand side is w.r.t. $D_n$, the first inequality follows from the definition of $G$, and the second inequality follows from Lemma \ref{lemma:node_sum}. Since by assumption $\ell^\prime \rightarrow \infty$ when $n \rightarrow \infty$, also $P\{V_{q(X)} > \delta, q(X) \in A, G\} \rightarrow 0$.

\emph{Term II:} Applying a similar technique as in \eqref{eq:second_term2}, we have
\[
%\label{eq:first_left_right3}    
\begin{split}
P\{q(X) \in B, G\} &= E  \left[\mathbbm{1}_G P\{q(X) \in B \,|\, D_n\} \right] \\
% &= E  \left[\mathbbm{1}_G \sum_{l=1}^{2^{\ell^\prime}} \mathbbm{1}\{l \in B\} \mu(R_l) \right]  \\ 
&= E  \left[\mathbbm{1}_G \sum_{l = 1}^{2^{\ell^\prime}} \mathbbm{1}\{l \in B\} \mu(R_l) \right]  \\ 
&\leq  \frac{1 + \epsilon}{2^{\ell^\prime}} E \left[  N_B \right] \\
&\leq \frac{1 + \epsilon}{2^{\ell^\prime}} \cdot 4d \cdot 2^{\ell^\prime - \frac{\ell^\prime}{d}}  \\
&=  4d \cdot \frac{1 + \epsilon}{2^{\ell^\prime / d}}, 
\end{split}
\]
where the expectation is w.r.t. $D_n$, the first inequality follows from the definition of $G$ and the second inequality follows from Lemma \ref{lemma:edge_bound}. Hence, also $P\{q(X) \in B, G\} \rightarrow 0$ when $n \rightarrow \infty$.

\emph{Term III:}  Finally, we bound the probability of the event $G^C$. Let $\mathcal{R}$ be the class of all hyperrectangles in $\mathbb{R}^d$. The Vapnik-Chervonenkis dimension of $\mathcal{R}$ is $2d$ (see, e.g., Theorem 13.8. by \citet[p. 220-221]{devroye2013probabilistic}), and hence we have $s(\mathcal{R},n) \leq n^{2d}$ for its shatter coefficient (see, e.g., Theorem 13.3. by \citet[p. 218]{devroye2013probabilistic}). If $n \geq 2 \cdot \frac{\log_2 n}{\epsilon} \geq 2 \cdot \frac{2^{\ell^\prime}}{\epsilon}$, then $\frac{1}{n} \leq \frac{1}{2} \cdot \frac{\epsilon}{2^{\ell^\prime}}$. This means that for large enough $n$, we have 
\begin{equation}
\label{eq:VC_bound}
\begin{split}
P(G^C) &= P\left\{\exists l \,\, \mathrm{s.t.} \,\, \mu(R_l) > \frac{1 + \epsilon}{2^{\ell^\prime}} \right\} \\
&= P\left\{\exists l \,\, \mathrm{s.t.} \,\, \mu(R_l) - \left(\frac{1}{2^{\ell^\prime}} + \frac{1}{n}\right) > \frac{\epsilon}{2^{\ell^\prime}} - \frac{1}{n} \right\} \\
&\leq P\left\{\exists l \,\, \mathrm{s.t.} \,\, \mu(R_l) - \mu_n(R_l) > \frac{\epsilon}{2^{\ell^\prime}} - \frac{1}{n}  \right\} \\
&\leq P\left\{\exists l \,\, \mathrm{s.t.} \,\, \mu(R_l) - \mu_n(R_l) > \frac{\epsilon}{2^{\ell^\prime + 1}}  \right\} \\
&\leq P\left\{\underset{R \in \mathcal{R}}{\sup} |\mu(R) - \mu_n(R)| > \frac{\epsilon}{2^{\ell^\prime + 1}} \right\} \\
&\leq 8 s(\mathcal{R},n) \exp \left\{ -\frac{n\epsilon^2}{128 \cdot 2^{2\ell^\prime}}\right\} \\
&\leq 8 n^{2d} \exp \left\{ -\frac{n\epsilon^2}{128 \cdot 2^{2\ell^\prime}}\right\} \\
&\leq 8 n^{2d} \exp \left\{ -\frac{n\epsilon^2}{128 (\log_2 n)^2}\right\} \rightarrow 0
\end{split}
\end{equation}
when $n \rightarrow \infty$.  The first inequality on the right hand side of \eqref{eq:VC_bound} follows because for the empirical measure---denoted by $\mu_n(A) := \frac{1}{n} \sum_{i=1}^n \mathbbm{1}_A(X_i)$ for any measurable set $A$---of any leaf $R_l$ it holds that $\mu_n(R_l) \leq \frac{1}{2^{\ell^\prime}} + \frac{1}{n}$. The fourth inequality follows from the Vapnik-Chervonenkis inequality \citep{vapnik1971uniform}; we use the version presented in Theorem 12.5. by \citet[p. 197-198]{devroye2013probabilistic}. The last inequality follows because $2^{2\ell^\prime} \leq (\log_2 n)^2$ by the definition of $\ell^\prime$.
\end{proof}

\section{Experimental setup}
\label{sec:experimental_setup}

All the algorithms, hyperparameter combinations, and code used in the experiments are included as supplementary material and can also be found at \url{https://github.com/vioshyvo/a-multilabel-classification-framework}.

\subsection{Computing environment}
\label{sec:computing_environment}

The experiments were ran on a machine with two Xeon E5-2680 v4 2.4GHz processors, 256GB RAM and CentOS 7 as the operating system. All queries were ran using only a single thread. The algorithms and test code were written in C++14 and compiled using GCC 5.4.0 with the optimization flags \texttt{-Ofast} and \texttt{-march=native}.

\subsection{Data sets}
\label{sec:data_sets}

Table \ref{table:datasets} contains the specifications of the data sets. We used four publicly available and commonly used benchmark data sets (Fashion\footnote{https://github.com/zalandoresearch/fashion-mnist}, GIST\footnote{http://corpus-texmex.irisa.fr}, STL-10\footnote{https://cs.stanford.edu/~acoates/stl10}, and Trevi\footnote{http://phototour.cs.washington.edu/patches}) consisting of raw or preprocessed images. We randomly divided the original data sets into the corpus, the validation set ($n_{\mathrm{validation}} = 1000$), and the test set ($n_{\mathrm{test}} = 1000$). The corpus $\{c_i\}_{i=1}^m$ was used as a training set. Since in the previous benchmarks for ANN search \citep{Aumuller2019a,Li2019} the problem is not considered in the machine learning setting, they do not use a distinct test set, but present the optimal results on the validation set. To follow this standard practice, we also present the results on the validation set, but note that the results were stable between the validation and test sets; there was some random variability, but we observed no signs of overfitting to the validation set.

\begin{table}[ht!]
\begin{center}
\caption{Data sets used in the experiments}
\label{table:datasets}
\begin{tabular}{l l l}
\toprule
Data set & corpus size $m$ & dimension $d$ \\
\midrule
Fashion & 58000 & 784 \\
GIST & 100000 & 960 \\
STL-10 & 98000 & 9216 \\
Trevi & 99120 & 4096 \\
\bottomrule 
\end{tabular}
\end{center}
\end{table}

\subsection{Hyperparameter settings}
\label{sec:hyperparameters}

According to our initial experiments, the performance of each type of a tree was robust w.r.t. its sparsity/randomization parameter (the number of the uniformly at random chosen coordinate directions $a \in \{1, \dots, d\}$ from which the optimal split direction was chosen for multiclass classification trees; the dimensionality $a$ of the random subspace in which first principal component was approximated for PCA trees; the expected number $a$ of the non-zero components in the random vectors onto which the node points are projected in RP trees; and the number $o$ of the highest variance directions of the node points from which the split direction was chosen uniformly at random in $k$-d trees). Therefore, we kept these parameters fixed in the final experiments: for multiclass classification trees and the PCA trees we used the value $a = \lceil\sqrt{d}\rceil$; for the RP trees the value $a=1/\sqrt{d}$; and for the $k$-d trees the value $o=5$. Further, we set the learning rate of the iterative PCA algorithm in PCA trees to $\gamma=0.01$ and the maximum number of iterations to $t=20$.

For the recall levels on the range [0.5, 0.99] considered in the article, the optimal numbers of trees $T$ were generally on the range [5, 200], the optimal depths of the trees were on the range [10, 15], and the optimal values of the threshold parameter $\tau$ were on the range [1, 20] for PCA, RP, and $k$-d trees that use  the raw counts as score function values. For multiclass classification trees that use the probability estimates \eqref{eq:conditional_probability_estimate} to select the candidate set, the optimal values of the threshold parameter $\tau$ were on the range [0.00001, 0.005]. We observed that using a value $k^\prime > k$ to learn the trees sometimes improved performance. We tested values $k^\prime \in\{10, 50, 100\}$ for learning the trees, with $k^\prime=10$ or $k^\prime=50$ usually being the optimal parameter value when $k=10$.  

For the other algorithms, we used the same hyperparameters as in ANN-benchmarks\footnote{https://github.com/erikbern/ann-benchmarks/blob/master/algos.yaml} as a starting point, and in many cases used even larger grids to ensure that the optimal hyperparameter settings were found. 

\section{Data structures}
\label{sec:data_structures}

In this section we review the five types of trees considered in this article (see \ref{sec:chronological_kd_tree}--\ref{sec:rpt} below).
%data structures (unsupervised PCA, RP, and $k$-d trees, and supervised multiclass classification trees) used in the experimental section. 
Random projection, PCA, $k$-d trees and chronological $k$-d trees have been widely used for ANN search (see, e.g., \citep{Silpa2008,Muja2014,Dasgupta2015,Jaasaari2019}), whereas the multiclass classification tree is a standard data structure for classification. For completeness, we include the full descriptions of the algorithms here. 

We begin by motivating the natural classifier \eqref{eq:natural_classifier} from the point of view of the multilabel problem reductions (see \citet{reddi2019stochastic,menon2019multilabel}). %This natural classifier can then be used in combination with 
First note that since the label set $L(X)$ is a deterministic function of the query point $X$---which means that the conditional label probabilities $\eta_1(x), \dots, \eta_m(x)$ are all equal to either 0 or 1---the Bayes classifier for 0-1 loss is obtained by thresholding these label probabilities by any $\tau \in [0,1)$. As a corollary, the same holds also for other less strict loss functions, such as precision, recall, and Hamming loss. This justifies following the common practice of estimating the conditional label probabilities by reducing the original multilabel classification problem into a series of binary or multiclass classification problems \citep{menon2019multilabel}. 

In the \emph{pick-all-labels} (PAL) \citep{reddi2019stochastic} reduction, a separate multiclass observation is created from each positive label, whereas in in the \emph{one-versus-all} (OVA) reduction, each of the $m$ labels is modeled as an independent binary classification problem. In the case of ANN search the maximum likelihood estimates for the label probabilities under the PAL and OVA reductions---i.e., under the multinomial and binomial models, respectively---are $\hat{\theta}_{\mathrm{PAL}} = \frac{1}{nk} \sum_{i=1}^n y_{ij}$ and $\hat{\theta}_{\mathrm{OVA}} = \frac{1}{n} \sum_{i=1}^n y_{ij}$. Since the size of the label set $L(x)$ is always $k$, the two estimates are proportional to each other, $\mathrm{\hat{\theta}_{\mathrm{OVA}}} = k\hat{\theta}_{\mathrm{PAL}}$, and hence, given the partition $\mathcal{P}$, the parameter estimates of the natural classifier \eqref{eq:natural_classifier}---i.e., observed label proportions among the training set points in a given partition element---minimise the log-likelihood under both reductions. % when the labels are modeled by the constant (multinomial/binomial) model(s) at each of the partition elements. 

Motivated by the above, we use the natural classifier \eqref{eq:natural_classifier} for prediction in combination with all tree types. When an ensemble of trees is used as a classifier, we compute the conditional label probability estimates of the ensemble as in \eqref{eq:natural_classifier_ensemble_og} by averaging the contributions of the individual trees.

\subsection{Chronological k-d tree}
\label{sec:chronological_kd_tree}

The chronological $k$-d tree \citep{Bentley1975} was the first data structure proposed for speeding up nearest neighbor search. It rotates the split directions and uses the same split direction for all the nodes at one level of a tree. At the first level the training data is split at the median of the first coordinates of the data points. At the second level both nodes are split at the median of the second coordinates of the node points. At the $(d+1)$th level, the nodes are split again at the median of the first coordinates, and so on (see Algorithm \ref{alg:KD-chronological-grow}).
More adaptive version of the $k$-d tree that splits at the coordinate direction in which the node points have the highest variance was proposed by \cite{Friedman1976}; we use a randomized version of this adaptive $k$-d tree (see Sec. \ref{sec:kd_tree} and Algorithm \ref{alg:KD-grow}) in the experiments of this article.

\begin{algorithm}
\begin{algorithmic}[1]
\State \textbf{Input:} a set of node points $X \subset \{x_1, \dots, x_n\}$, current level $\ell^\prime$, maximum height $\ell$ 
\State \textbf{Output:} a node of a tree
\Procedure{grow-kd}{$X$, $\ell^\prime$, $\ell$}
  \If{$\ell^\prime =\ell$}
    \State \Return{$X$ node as a leaf node}
  \EndIf
  \State $\hat{r} \gets (\ell^\prime \,\, \mathrm{modulo} \,\, d) + 1$ 
  \State $\hat{s} \gets$ median of the $\hat{r}$th coordinates of the node points
  \State $\mathrm{left} \gets$ \Call{grow-kd}{$\{x_i \in X \,:\, x_{i\hat{r}} \leq \hat{s}\}$, $\ell^\prime + 1$, $\ell$}
  \State $\mathrm{right} \gets$ \Call{grow-kd}{$\{x_i \in X \,:\, x_{i\hat{r}} > \hat{s}\}$, $\ell^\prime + 1$, $\ell$}
  \State \Return{$(\mathrm{left}, \mathrm{right}, \hat{r}, \hat{s}$)} as an inner node
\EndProcedure
\end{algorithmic}
\caption{Grow a chronological $k$-d tree \citep{Bentley1975}}
\label{alg:KD-chronological-grow}
\end{algorithm}

\subsection{Multiclass classification tree}
\label{sec:supervised_classification_tree}

As discussed at the beginning of the section, the maximum likelihood estimates of the piecewise constant model under both the PAL and OVA reductions coincide in the special case of ANN search. Thus, in principle it would make no difference which one of these reductions we employed to learn the classification trees. However, in practice computation of the binomial likelihood requires keeping track of the contributions of the negative labels which is inconvenient when the label space is large. Thus, we employ the PAL reduction where each positive label is modeled by a separate multiclass observation, and learn the standard multiclass classification trees by greedily maximising the multinomial log-likelihood (i.e., use the multiclass cross-entropy as a split criterion); see Algorithm \ref{alg:RF-grow} for details.

To grow a random forest, we randomize the multiclass classification trees by optimising the split point only in randomly chosen $a=\lceil \sqrt{d}\rceil$ coordinate directions at each node of a tree. We do not use bootstrap samples, but fit each tree to the original training data. To decrease learning time, we subsample 100 training points at each node (if node size $>100$), and use only this subset to optimize the splits; we did not observe any negative impact on prediction performance.

\begin{algorithm}[tb!]
\begin{algorithmic}[1]
\State \textbf{Input:} a set of node points $X \subset \{\ve{x}_1, \dots, \ve{x}_n\}$, current depth, maximum depth $\ell$, sparsity parameter $a$ 
\State \textbf{Output:} a node of a tree
\Procedure{grow}{$X$, depth, $\ell, a = \lceil \sqrt{d} \rceil$}
  \If{depth $=\ell$}
    \State \Return{$X$ as a leaf node}
  \EndIf
  \State $D \gets$ $a$ random unique dimensions from $\{1, \dots, d\}$
  \State $(\mathrm{maxgain}, \hat{r}, \hat{s}) \gets (0, 0, 0)$
  \State $N \gets |X|$
  \For{$r \in D$}
    \State let $s_1 \leq s_2 \leq \dots \leq s_N$ be the $r$th coordinate of points $x \in X$ sorted in ascending order
    \For{$s \in \{s_1, \dots, s_N\}$}
    \State $X_{\mathrm{left}}\gets \{\ve{x}_i \in X \,:\, \ve{x}_{ir} \leq s  \}$; $\,\,N_{\mathrm{left}} = |X_{\mathrm{left}}|$
    \State $X_{\mathrm{right}} \gets \{\ve{x}_i \in X \,:\, \ve{x}_{ir} > s \}$;$\,\,N_{\mathrm{right}} = |X_{\mathrm{right}}|$
    \For{$j \in \{1, \dots, m\}$}
    \State $v_j^{(\mathrm{left})} \gets \sum_{\ve{x}_i \in X_{\mathrm{left}}} y_{ij}$
    \State $v_j^{(\mathrm{right})} \gets \sum_{\ve{x}_i \in X_{\mathrm{right}}} y_{ij}$
    \State $v_j \gets \sum_{\ve{x}_i \in X} y_{ij}$
    \State  $\hat{\parameter}_j^{(\mathrm{left})} := v_j^{(\mathrm{left})} / (kN_{\mathrm{left}})$ 
    \State $\hat{\parameter}_j^{(\mathrm{right})} := v_j^{(\mathrm{right})} / (kN_{\mathrm{right}})$
    \State $\hat{\parameter}_j := v_j / (kN)$
    \EndFor
    \State $\mathrm{gain} \gets \sum_{j=1}^m v_j^{(\mathrm{left})} \log \hat{\parameter}_j^{(\mathrm{left})} + \sum_{j=1}^m v_j^{(\mathrm{right})} \log \hat{\parameter}_j^{(\mathrm{right})} - \sum_{j=1}^m v_j \log \hat{\parameter}_j$
    \If{$\mathrm{gain} > \mathrm{maxgain}$}
      \State $(\mathrm{maxgain}, \hat{r}, \hat{s}) \gets (\mathrm{gain}, r, s)$
    \EndIf
    \EndFor
  \EndFor
  \If{$\mathrm{maxgain} \leq 0$}
    \State \Return{$X$ as a leaf node}
  \EndIf
  \State $\mathrm{left} \gets$ \Call{grow}{$\{\ve{x}_i \in X \,:\, \ve{x}_{i\hat{r}} \leq \hat{s}\}$, depth + 1, $\ell, a$}
  \State $\mathrm{right} \gets$ \Call{grow}{$\{\ve{x}_i \in X \,:\, \ve{x}_{i\hat{r}} > \hat{s}\}$, depth + 1 $\ell, a$}
  \State \Return{$(\mathrm{left}, \mathrm{right}, \hat{r}, \hat{s})$} as an inner node
\EndProcedure
\end{algorithmic}
\caption{Grow a randomized multiclass classification tree}
\label{alg:RF-grow}
\end{algorithm}

\subsection{k-d tree}
\label{sec:kd_tree}

Algorithm~\ref{alg:KD-grow} details the recursive algorithm for growing a randomized $k$-d tree. As in multiclass classification trees, the splits are restricted to the directions of the coordinate axes. In $k$-d trees~\citep{Friedman1976} the normal of the splitting hyperplane is chosen as the coordinate direction $r \in \{1, \dots, d\}$ along which the node points have the highest variance. The split point $\hat{s}$ is chosen as the median of the $\hat{r}$th coordinate of the node points. To grow an ensemble of randomized trees, we use the randomization scheme proposed by \cite{Silpa2008}: instead of splitting at the direction of the highest variance, we draw uniformly at random one of the $o$ highest variance directions and use it as a split direction $\hat{r}$. We use the default value $o=5$ recommended by~\cite{Muja2014} for this hyperparameter.

\begin{algorithm}[tb!]
\begin{algorithmic}[1]
\State \textbf{Input:} a set of node points $X \subset \{\ve{x}_1, \dots, \ve{x}_n\}$, current level, maximum height $\ell$, the number of highest variances directions from which the split direction is sampled $o$ 
\State \textbf{Output:} a node of a tree
\Procedure{grow-kd}{$X$, level, $\ell, o = 5$}
  \If{level $=\ell$}
    \State \Return{$X$ as a leaf node}
  \EndIf
  \State $D \gets$ set of $o$ coordinate directions in which the node points have the highest variance
  \State $\hat{r} \gets$ uniformly at random sampled dimension from the set $D$
  \State $\hat{s} \gets$ median of $\hat{r}$th coordinates of the node points
  \State $\mathrm{left} \gets$ \Call{grow-kd}{$\{\ve{x}_i \in X \,:\, \ve{x}_{i\hat{r}} \leq \hat{s}\}$, level + 1, $\ell, o$}
  \State $\mathrm{right} \gets$ \Call{grow-kd}{$\{\ve{x}_i \in X \,:\, \ve{x}_{i\hat{r}} > \hat{s}\}$, level + 1 $\ell, o$}
  \State \Return{$(\mathrm{left}, \mathrm{right}, \hat{r}, \hat{s}$)} as an inner node
\EndProcedure
\end{algorithmic}
\caption{Grow a randomized $k$-d tree \citep{Silpa2008}}
\label{alg:KD-grow}
\end{algorithm}

\subsection{PCA tree}

In a PCA tree~\citep{Sproull1991}, the projection direction at each node is the first principal component, i.e. the direction the node points have the greatest variance when projected onto. PCA trees are on the one hand slow to compute, as computing exact PCA is expensive, and on the other hand they are deterministic and thus multiple trees cannot be grown to boost accuracy.

To solve the first problem, ~\cite{mccartin2012approximate} proposed an \emph{approximate PCA tree}, which uses gradient descent updates to approximate the first principal component of the data at each node of the tree. To address the second problem, \cite{Jaasaari2019} proposed a \emph{sparse approximate PCA tree}, which draws at each node of the tree uniformly at random only $a = \sqrt{d}$ dimensions, and computes the approximate first principal component in the subspace defined by these dimensions.

The gradient descent update for approximate PCA is
\[
r_t := r_{t-1} + \gamma\,  \text{Cov}(Z) r_{t - 1}, \qquad r_t := r_t / \left\| r_t \right\|_2,
\]
where $r_t$ is the projection vector at time $t$, $Z$ is a matrix containing the data, and $\gamma$ is the learning rate which we fix as $0.01$. We did not observe further tuning of this hyperparameter to be necessary.

Algorithm~\ref{alg:PCA-grow} details a recursive algorithm for growing a sparse approximate PCA tree. Algorithm~\ref{alg:PCA} details the actual approximate PCA algorithm used to find the split direction. On line 7, the matrix $Z$ is formed by taking the vectors of the current points $X$ as columns of a matrix and then slicing only the rows (dimensions) that were randomly selected into the set $D$ on line 6. The sample covariance matrix $C$ is formed from $Z$ on lines 8-9. Lines 10-11 initialize the projection vector from the unit sphere, while lines 12-17 implement the gradient descent update described above. By default, we do $t = 20$ iterations of gradient descent, unless the $\ell_1$ norm of the projection vector changes by less than $\epsilon = 0.01$ in a single iteration.

\begin{algorithm}
\begin{algorithmic}[1]
\State \textbf{Input:} a set of node points $X \subset \{\ve{x}_1, \dots, \ve{x}_n\}$, current depth, maximum depth $\ell$, sparsity parameter $a$, maximum number of iterations $t$, learning rate $\gamma$, threshold parameter $\epsilon$ 
\State \textbf{Output:} a node of a tree
\Procedure{grow}{$X$, depth, $\ell, a = \lceil \sqrt{d} \rceil, t = 20, \gamma = 0.01, \epsilon = 0.01$}
  \If{depth $=\ell$}
    \State \Return{$X$ as a leaf node}
  \EndIf
    \State direction $\gets$ \Call{pca-generate-split-direction}{$X, a, t, \gamma, \epsilon$}
  \State proj $\gets$ \Call{project}{$X$, direction}
  \State $\hat{s} \gets$ \Call{median}{proj}
  \State $X_\mathrm{left} \gets$ points in $X$ for which $\mathrm{proj} \leq \hat{s}$
  \State $X_\mathrm{right} \gets$ points in $X$ for which $\mathrm{proj} > \hat{s}$
  \State left $\gets$ \Call{grow}{$X_\mathrm{left}$, depth + 1, $\ell, a, t, \gamma, \epsilon$}
  \State right $\gets$ \Call{grow}{$X_\mathrm{right}$, depth + 1, $\ell, a, t, \gamma, \epsilon$}
  \State \Return{(left, right, direction, $\hat{s}$)} as an inner node
\EndProcedure
\end{algorithmic}
\caption{Grow a randomized PCA tree \citep{Jaasaari2019}}
\label{alg:PCA-grow}
\end{algorithm}

\begin{algorithm}
\begin{algorithmic}[1]
\State \textbf{Input:} a set of node points $X \subset \{\ve{x}_1, \dots, \ve{x}_n\}$, sparsity parameter $a$, maximum number of iterations $t$, learning rate $\gamma$, threshold parameter $\epsilon$ 
\State \textbf{Output:} an approximate first principal component $\ve{p}$ in $a$-dimensional subspace; observe that $\ve{p}$ has only $a$ nonzero components.
\Procedure{pca-generate-split-direction}{$X, a, t, \gamma, \epsilon$}
  \State $N \gets |X|$
  \State initialize $d$-dimensional vector $\ve{p}$ with zeros
  \State $D \gets$ $a$ random unique dimensions from $1, \dots, d$
  \State $\ve{Z} \gets$ points in $X$ with all components but those in $D$ removed
  \State $\ve{M} \gets \ve{Z}(\ve{I}_{N}-{\tfrac {1}{N}}\mathbf {1}
  \mathbf{1^{\text{T}}})\ve{Z}^T$
  \State $\ve{C} \gets \frac{1}{N - 1} \ve{M}\ve{M}^T$
  \State $\ve{r} \gets \mathcal{X}_a({\bf 0}, {\bf I})$
  \State $\ve{r} \gets \ve{r} / \|\ve{r}\|_2$
  \For{$i \in \{1, \dots, t\}$}
    \State $\ve{r}' \gets \ve{r}$
    \State $\ve{r} \gets \ve{r} + \gamma\ve{C}\ve{r}$
    \State $\ve{r} \gets \ve{r} / \|\ve{r}\|_2$
    \If{$\|\ve{r} - \ve{r}'\|_1 < \epsilon$}
      \State break
    \EndIf
  \EndFor
  \State $j \gets 1$
  \For{$i \in D$}
   \State $\ve{p}[i] \gets \ve{r}[j]$
   \State $j \gets j + 1$
  \EndFor
  \State \Return{$\ve{p}$}
\EndProcedure
\end{algorithmic}
\caption{Generate projection vector for a randomized PCA tree \citep{Jaasaari2019}}
\label{alg:PCA}
\end{algorithm}

\subsection{Random projection tree}
\label{sec:rpt}

Algorithm~\ref{alg:RP-grow} describes a process of growing a sparse random projection (RP) tree. In RP trees \citep{Dasgupta2008,Dasgupta2015}, the normal of the splitting hyperplane is chosen uniformly at random from $d$-dimensional standard normal distribution $N(\ve{0}, \ve{I})$. The node points are projected into this random vector, and the node is split at the median of the projections. \cite{Hyvonen2016} proposed a sparse variant, where the components of the random vectors are generated from the standard normal distribution with the probability $a$, and are zero with the probability $1-a$ (see Algorithm~\ref{alg:RP}). For this hyperparameter we use the default value $a = 1/\sqrt{d}$ as recommended by \cite{Hyvonen2016}. This decreases both the index construction time and the query routing time by a factor of $\sqrt{d}$ compared to the original dense RP trees.

\begin{algorithm}
\begin{algorithmic}[1]
\State \textbf{Input:} set of node points $X \subset \{\ve{x}_1, \dots, \ve{x}_n\}$, current depth, maximum depth $\ell$, sparsity parameter $a$ 
\State \textbf{Output:} node of a tree
\Procedure{grow-rp}{$X$, depth, $\ell, a = 1/\sqrt{d}$}
  \If{depth $=\ell$}
    \State \Return{$X$ as a leaf node}
  \EndIf
    \State direction $\gets$ \Call{rp-generate-split-direction}{$a$}
  \State proj $\gets$ \Call{project}{$X$, direction}
  \State $\hat{s} \gets$ \Call{median}{proj}
  \State $X_\mathrm{left} \gets$ points in $X$ for which $\mathrm{proj} \leq \hat{s}$
  \State $X_\mathrm{right} \gets$ points in $X$ for which $\mathrm{proj} > \hat{s}$
  \State $\mathrm{left} \gets$ \Call{grow-rp}{$X_\mathrm{left}$, depth + 1, $\ell, a$}
  \State $\mathrm{right} \gets$ \Call{grow-rp}{$X_\mathrm{right}$, depth + 1, $\ell, a$}
  \State \Return{$(\mathrm{left}, \mathrm{right}, \mathrm{direction}, \hat{s})$} as an inner node
\EndProcedure
\end{algorithmic}
\caption{Grow a sparse RP tree \citep{Hyvonen2016}}
\label{alg:RP-grow}
\end{algorithm}

\begin{algorithm}
\begin{algorithmic}[1]
\State \textbf{Input:} sparsity parameter $a$ that is the expected proportion of non-zero components in the output vector
\State \textbf{Output:} a random $d$-dimensional vector 
\Procedure{rp-generate-split-direction}{$a$}
  \State initialize $d$-dimensional vector $p$ with zeros
  \For{$i \in \{1, \dots, d\}$}
    \If{\Call{random}{0, 1} $\le a$}
      \State generate $p[i] \sim N(0,1)$
    \EndIf
  \EndFor
  \State \Return{$p$}
\EndProcedure
\end{algorithmic}
\caption{Generate a normal of the splitting hyperplane for a sparse RP tree \citep{Hyvonen2016}}
\label{alg:RP}
\end{algorithm}

\section{Additional experimental results}
\label{sec:additional_experimental_results}

\subsection{Index construction times}
\label{sec:index_construction_times}

The index construction times of the algorithms at the optimal parameters for the recall levels $R = 0.8, 0.9, 0.95$ are shown in Table \ref{table:build_times}. The computation time for finding the nearest neighbors (i.e., the labels) of the corpus points is not included in the index construction times, since they are computed only once for each data set; they are found in Table \ref{table:knn-times} (in the column "exact").

The ensembles of unsupervised trees are relatively fast to build, especially on the high-dimensional data sets: on STL-10, PCA trees have the fastest query times, and are an order of magnitude faster to train compared to the graph and quantization methods. 

The multiclass classification trees (RF) are slower to train compared to the unsupervised (PCA, KD, and RP) trees. We expect that their training times could be decreased by standard techniques, such as using weighted quantile sketches~\citep{greenwald2001space,zhang2007fast,Chen2016} when optimizing the split points.

\begin{table}[!hbtp]
\begin{center}
\caption{Index building time (seconds) at optimal parameters. The fastest time for each recall level is typeset in bold.}
\label{table:build_times}
\begin{tabular}{c c c c c c | c c c }
\toprule
data set & R (\%) & PCA & KD & RP & RF & ANNOY & HNSW & IVF-PQ \\
\midrule
 & 80 & 2.014 & {\bf 0.929} & 1.298 & 14.422 & 1.244 & 1.518 & 5.867 \\
Fashion & 90 & 1.621 & 1.500 & {\bf 1.284} & 25.591 & 11.564 & 1.690 & 5.867 \\
 & 95 & 1.847 & 2.208 & 1.925 & 45.683 & 11.564 & {\bf 1.518} & 5.867 \\
\midrule
 & 80 & 27.732 & 27.162 & 30.520 & 131.430 & 16.349 & 19.114 & {\bf 13.031} \\
GIST & 90 & 30.313 & 36.664 & 56.624 & 300.340 & 62.931 & 21.560 & {\bf 13.031} \\
 & 95 & 30.313 & 49.707 & 48.056 & 300.340 & {\bf 8.319} & 26.456 & 13.031 \\
\midrule
 & 80 & {\bf 4.497} & 25.426 & 12.204 & 316.790 & 32.036 & 93.393 & 92.577 \\
STL-10 & 90 & {\bf 8.891} & 30.814 & 12.145 & 647.320 & 489.482 & 132.500 & 92.577 \\
 & 95 & {\bf 7.918} & 28.286 & 12.145 & 466.520 & 489.480 & 201.070 & 92.577 \\
\midrule
 & 80 & {\bf 4.900} & 11.158 & 10.185 & 420.250 & 141.794 & 60.044 & 43.520 \\
Trevi & 90 & 18.937 & 13.886 & {\bf 10.169} & 420.250 & 141.790 & 60.044 & 43.520 \\
 & 95 & 18.937 & 12.432 & {\bf 11.261} & 420.250 & 141.790 & 60.044 & 43.520 \\
\bottomrule
\end{tabular}
\end{center}
\end{table}

\subsection{Comparison to graph and quantization methods}
\label{sec:experiments_graphs_quantization}

To empirically justify studying partition-based ANN algorithms, we also include in the comparison Hierarchical Navigable Small World (HNSW) \citep{Malkov2018} graphs and Inverted File Product Quantization (IVF-PQ) \cite{Jegou2010}, that were the fastest graph-based and the fastest quantization-based algorithm, respectively, according to ANN-benchmarks~\citep{Aumuller2019a} project at the time of its publication\footnote{As of May 2022, the fastest graph-based method is NGTQG (\url{https://github.com/yahoojapan/NGT/blob/master/bin/ngtqg/README.md}), the fastest quantization-based algorithm is SCANN \citep{guo2020accelerating} (see \url{http://ann-benchmarks.com/index.html} for updated results).}. For completeness, we also include the commonly-used tree-based method ANNOY\footnote{\url{https://github.com/spotify/annoy}} in the comparison. See Table \ref{table:comparison} for the results. We emphasize that this is not a benchmark article with the goal of proposing a single ANN algorithm and demonstrating its superiority over the competition---rather, we aim to establish a widely applicable theoretical framework for partition-based ANN search.

\begin{table*}[hbtp]
\begin{center}
\caption{Query times (seconds / 1000 queries) at different recall levels for the different tree types. The fastest method in each case is typeset in boldface.}
\label{table:comparison_graphs_quantization}
\begin{tabular}{l c c c c c | c c c }
\toprule
data set & R (\%) & PCA & KD & RP & RF & ANNOY & HNSW & IVF-PQ \\
\midrule
 & 80 & 0.075 & 0.076 & 0.099 & {\bf 0.063} & 0.193 & 0.064 & 0.266 \\
Fashion & 90 & 0.111 & 0.126 & 0.172 & 0.095 & 0.296 & {\bf 0.089} & 0.291 \\
 & 95 & 0.163 & 0.171 & 0.261 & 0.146 & 0.419 & {\bf 0.097} & 0.340 \\
\midrule
 & 80 & 1.330 & 0.958 & 1.009 & 0.705 & 2.525 & {\bf 0.524} & 0.872 \\
GIST & 90 & 2.942 & 2.286 & 2.226 & 1.530 & 5.973 & {\bf 0.819} & 2.037 \\
 & 95 & 5.641 & 4.451 & 4.598 & 3.253 & 7.477 & {\bf 1.212} & 2.657 \\
\midrule
 & 80 & {\bf 0.382} & 0.872 & 1.211 & 0.756 & 21.110 & 1.473 & 6.140 \\
STL-10 & 90 & {\bf 0.756} & 2.126 & 3.248 & 1.774 & 24.826 & 2.717 & 6.860 \\
 & 95 & {\bf 1.315} & 4.376 & 7.330 & 3.654 & 35.459 & 3.963 & 6.860 \\
\midrule
 & 80 & {\bf 0.330} & 0.543 & 0.591 & 0.582 & 5.259 & 0.705 & 1.677 \\
Trevi & 90 & {\bf 0.684} & 1.464 & 1.468 & 1.234 & 9.921 & 1.202 & 1.892 \\
 & 95 & {\bf 1.212} & 3.244 & 3.289 & 2.350 & 17.172 & 1.896 & 2.655 \\
\bottomrule
\end{tabular}
\end{center}
\end{table*}

\subsection{Comparison of candidate set selection methods: all data sets}

\begin{figure*}[tb!]
\centering
\includegraphics[width=\textwidth]{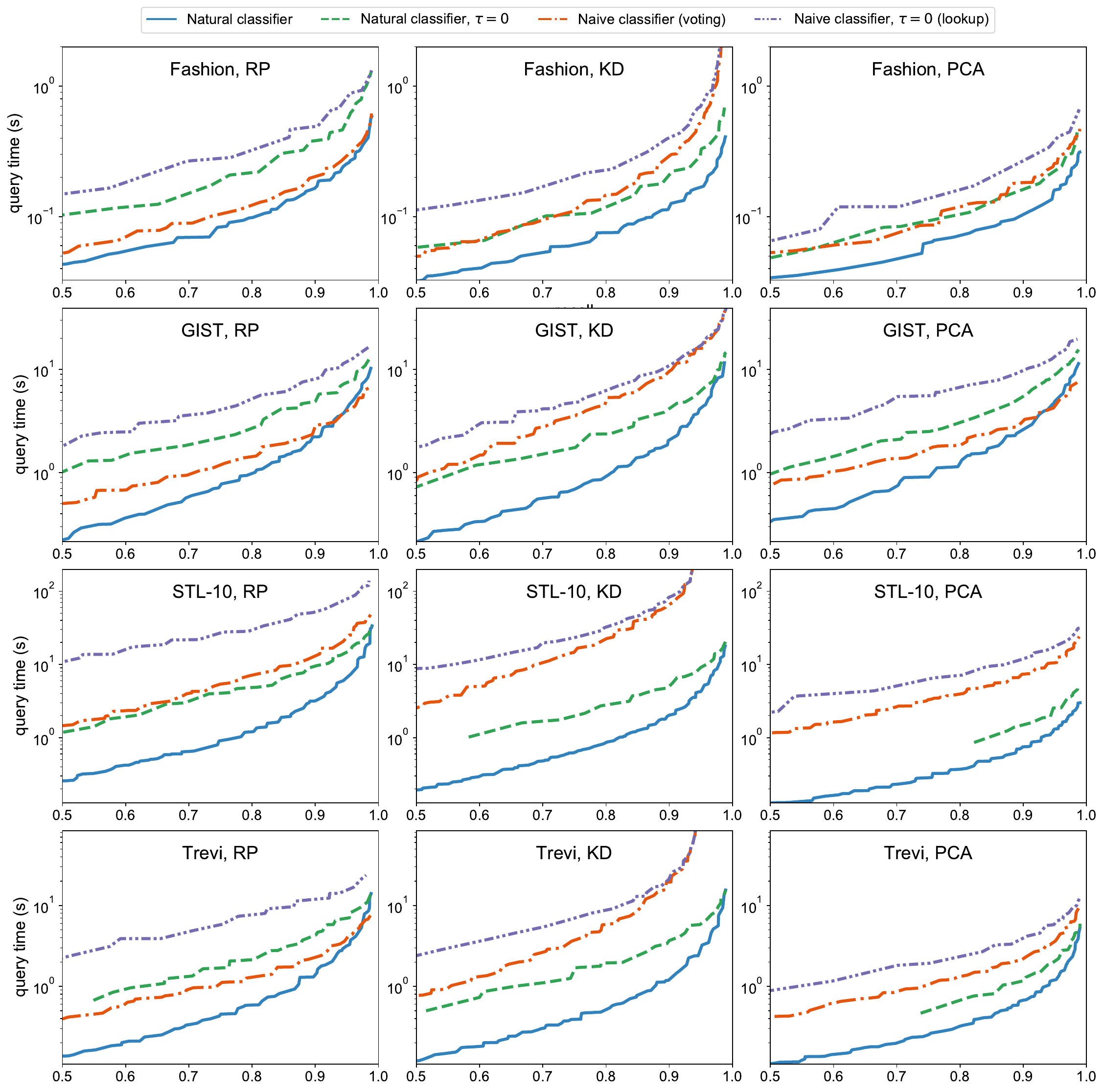}
\caption{Recall vs. query time (log scale) of ensembles of, RP, $k$-d, and PCA trees. The solid blue line is the natural classifier proposed in this paper; the dash-dotted red line is the natural classifier with $\tau = 0$ that is included for completeness; the dashed green line is voting; and the double-dash-dotted violet line is lookup search. The natural classifier is the fastest and the lookup search is the slowest of the methods for each tree type.}
\label{fig:search_methods_full}
\end{figure*}

Figure \ref{fig:search_methods_full} contains the results of the comparison between the candidate set selection for all the four data sets. The results validate the theoretical findings: using space-partitioning trees as natural classifiers to select the candidate set  improves their performance consistently for all the tree types and data sets compared to retrieving the candidate set in the lookup-based paradigm (lookup search and voting). Voting also always outperforms lookup search---this is not surprising since lookup search is a special case of voting with $\tau=0$ as discussed in Sec. \ref{sec:related_work}. For completeness, we also include the corresponding special case of the natural classifier with $\tau=0$ in the comparison.

\subsection{Training classifiers with noisy labels}
\label{sec:noisy-labels}

The disadvantage of the supervised ANN search algorithms compared to the purely unsupervised algorithms is that they require computing the true nearest neighbors $\{\ve{y}_i\}_{i=1}^n$ of the training set points $\{\ve{x}_i\}_{i=1}^n$, which is an $\mathcal{O}(nmd)$ operation. This is not a problem for the benchmark data sets used in this article---for the largest data set (STL-10, $n=98000$, $m=98000$, $d=9216$), computing the ground truth took 50 minutes on a single machine---but in the typical applications of ANN search the corpus size may be hundreds of millions or even billions.

The labels used to train the classifier do not have to be exact. We can also compute the \emph{approximate} nearest neighbors of the training set $\{\ve{x}_i\}_{i=1}^m$ and use them as labels $\{\ve{y}_i\}_{i=1}^m$ to train the classifier. For instance, using approximate nearest neighbors computed at average recall level of 90\% amounts to using noisy labels with 10\% noise. 

To test how the noisy labels affect the performance, we fit the random forest to training sets with 10\%, 20\%, 30\%, 40\%, and 50\% label noise. The noisy labels are obtained by running
the MRPT algorithm~\citep{Hyvonen2016} (i.e., an ensemble of RP trees where the candidate set is selected by voting) in combination with the automatic hyperparameter tuning algorithm~\citep{Jaasaari2019} to find the approximate nearest neighbors of the training set points with recall levels 
90\%, 80\%, 70\%, 60\%, and 50\%, respectively.

\begin{figure*}[tb!]
\centering
\includegraphics[width=.6\textwidth]{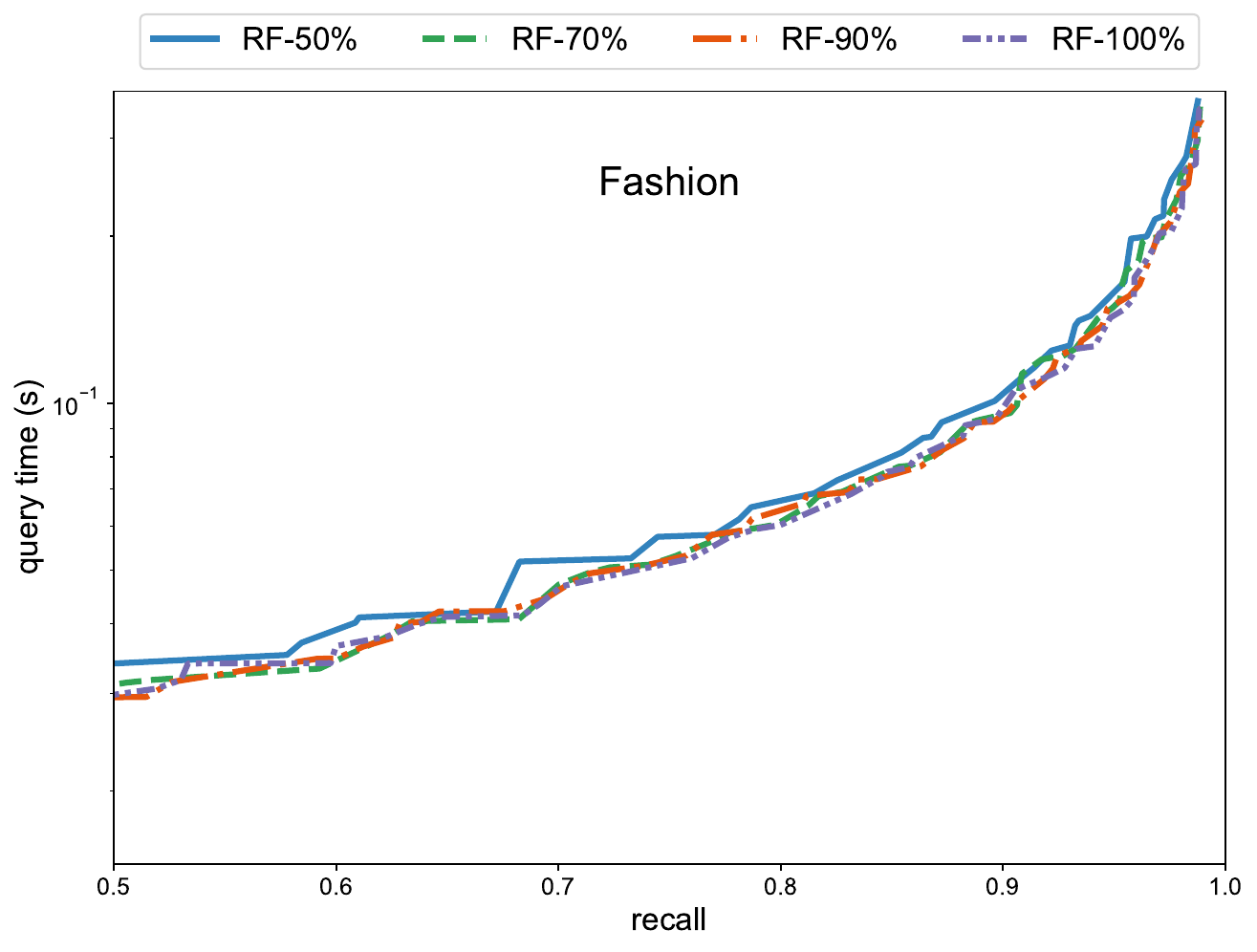}
\caption{Recall vs. query time (log scale) of a random forest trained with different amounts of label noise on the Fashion data set. RF-100\% is the random forest trained using the exact $k$-nn matrix, RF-90\% is the random forest trained using approximate $k$-nn matrix with that contains on average 90\% of the correct neighbors, etc. Allowing 10\% noise in labels has no visible effect on performance, and even allowing 50\% noise has very little effect.}
\label{fig:ann-fashion}
\end{figure*}

The results (c.f. Fig \ref{fig:ann-fashion}) indicate that tree-based classifiers are robust with respect to label noise: training the random forest with 10\% label noise has no visible effect on the performance of the algorithm, and even training on labels with 50\% noise has very little effect.

Computing times for exact and approximate nearest neighbors for the training set are found on Table \ref{table:knn-times} for all the four data sets. The results indicate that significant savings in preprocessing time can be obtained by using noisy labels: for instance, on STL-10 computing the exact computation took 50 minutes, whereas the approximate computation took only 1-10 minutes depending on the recall level.

\begin{table}[tb!]
\begin{center}
\caption{Computation times for exact (brute force) and noisy (MRPT algorithm) labels in seconds. The percentage is the average number of correct approximate nearest neighbors% in the $k$-nn matrix
.}
\label{table:knn-times}
\begin{tabular}{l l l l l l}
\toprule
data set & exact & 95\% & 90\% & 70\% & 50\% \\
\midrule
Fashion & 105.8 & 5.6 & 3.1 & 1.8 & 1.0 \\
GIST & 371.7 & 92.7 & 61.9 & 20.2 & 10.0  \\
Trevi & 1450.2 & 118.6 & 104.4 & 31.4 & 24.1  \\
STL-10 & 2992.7 & 596.3 & 409.0 & 97.2 & 66.5 \\
\bottomrule
\end{tabular}
\end{center}
\end{table}

\end{document}